%% file: main.tex
\documentclass{article}

\usepackage{iclr2027_conference,times}

\input{math_commands.tex}

\usepackage{hyperref}
\usepackage{url}
\usepackage{graphicx}
\usepackage{array}
\usepackage{tabularx}
\usepackage{subcaption}
\usepackage{float}
\usepackage{algorithm}
\usepackage{algorithmic}

\AtBeginDocument{
  \setlength{\textfloatsep}{10pt plus 2pt minus 2pt}
  \setlength{\floatsep}{8pt plus 2pt minus 2pt}
  \setlength{\intextsep}{8pt plus 2pt minus 2pt}
  \setlength{\abovedisplayskip}{6pt plus 2pt minus 2pt}
  \setlength{\belowdisplayskip}{6pt plus 2pt minus 2pt}
  \setlength{\abovedisplayshortskip}{3pt plus 2pt}
  \setlength{\belowdisplayshortskip}{4pt plus 2pt minus 2pt}
  \setlength{\abovecaptionskip}{4pt}
  \setlength{\belowcaptionskip}{4pt}
}

\title{Lightweight and Versatile Learned Optimization by Recombination of Gradient History}

\author{
    Minyoung Choi \quad Dalta Imam Maulana \quad Wanyeong Jung \\
    KAIST \\
    Daejeon, Republic of Korea \\
    \texttt{\{myconejo, daltamaulana, wanyeong\}@kaist.ac.kr}
}
\iclrfinalcopy

\begin{document}
\raggedbottom

\maketitle

\lhead{Preprint}

\input{sections/0_abstract}
\input{sections/1_introduction}
\input{sections/2_related_works}
\input{sections/3_method_policy}
\input{sections/4_evaluation}
\input{sections/6_conclusion}

\bibliography{iclr2027_conference}
\bibliographystyle{iclr2027_conference}

\clearpage
\appendix
\input{sections/A_evaluation_details}
\input{sections/B_additional_ablations}
\input{sections/C_structural_extension}

\end{document}

%% file: math_commands.tex
\usepackage{amsmath,amsfonts,bm}

\def\eqref#1{equation~\ref{#1}}

\def\1{\bm{1}}

\DeclareMathAlphabet{\mathsfit}{\encodingdefault}{\sfdefault}{m}{sl}
\SetMathAlphabet{\mathsfit}{bold}{\encodingdefault}{\sfdefault}{bx}{n}



%% file: sections/0_abstract.tex
\begin{abstract}
This paper presents a lightweight and versatile learned optimizer that dynamically recombines gradient history, represented as averages over disjoint time spans. 
The optimizer reduces the prediction space to one scalar coefficient  per gradient average, shared by multiple parameters.
Progressively averaging older gradients minimizes memory cost of long history, while keeping their contributions independently accessible.
A 37k-parameter network trained in 0.87 GPU-hours generalizes zero-shot to unseen tasks, lowering validation loss by 8.4\% and 0.1\% on BERT-Tiny and GPT-Tiny with embeddings and output head trained with Adam, and improving test accuracy over Adam by 5.8 \%p on a Vision Transformer and by 2.7 \%p on average across nine graph models.
\end{abstract}

%% file: sections/1_introduction.tex
\section{Introduction}

\begin{figure}[h]
    \centering
    \includegraphics[width=0.57\linewidth]{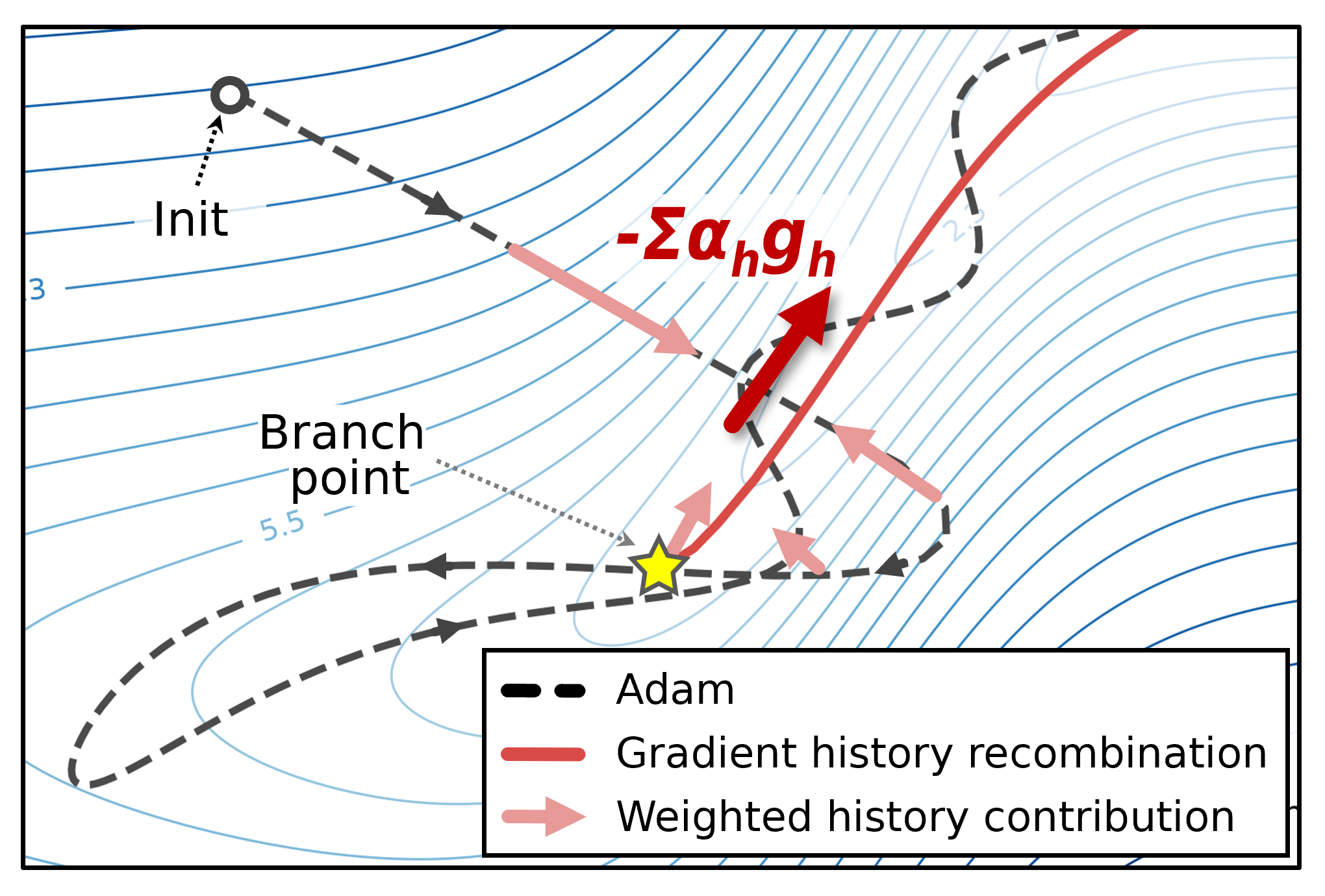}
    \caption{
        \textbf{Recombination of gradient history as a learned update.}
        Recombination of history contributions \(-\sum_h \alpha_h g_h\) with predicted coefficients \(\alpha_h\) can redirect the Adam trajectory away from ascent.
    }
    \label{fig:temporal-motivation}
\end{figure}

While learned optimizers can adapt to loss landscapes, designing lightweight and versatile architectures remains challenging.
Coordinate-wise prediction methods \citep{andrychowicz2016learning,wichrowska2017learned,metz2022velo} are expensive as they process gradient-derived features for individual parameters.
For example, training VeLO requires approximately 4,000 TPU-months \citep{metz2022velo}, more than twice the compute to train large language models \citep{chowdhery2023palm}.

Gradient history can serve directly as a useful directional basis for optimizers, as
demonstrated by momentum and Adam \citep{polyak1964some,kingma2014adam}.
This motivates an adaptive optimizer that learns to dynamically recombine gradients from the past optimization trajectory.

Building on this principle, this paper presents a learned optimizer that dynamically recombines gradient history, represented by averages over disjoint time spans.
To remain lightweight, the optimizer operates on compact statistics at coarse, configurable parameter granularity.
This also allows the optimizer to jointly compare different history components and adjust their relative contributions, while the disjoint representation keeps older time spans independently accessible.
Figure~\ref{fig:temporal-motivation} visualizes this concept, showing how past gradients are weighted and recombined into each update.

Two techniques make the optimizer versatile and practical.
First, fixed-dimensional representation of group statistics supports arbitrary granularity as well as versatility across target networks.
Second, progressive averaging of older gradients limits memory growth logarithmically with history length.

The core contributions are:
\begin{itemize}
    \item The presented learned optimizer justifies the use of gradient-history directions, jointly weighting and dynamically recombining them to construct each update.
    \item A single trained policy generalizes zero-shot to unseen language, vision, and graph benchmarks, improving validation loss and test accuracy over Adam.
    \item A compact 37k-parameter policy is trained in under one GPU-hour on an NVIDIA B200.
\end{itemize}

%% file: sections/2_related_works.tex
\section{Related Work}

\paragraph{Recombination of gradient history.}
The presented optimizer dynamically assigns contributions to gradient averages over disjoint time ranges without further task-specific tuning.
In contrast, typical optimizers that combine gradient history, such as AggMo and AdEMAMix~\citep{lucas2018aggregated,pagliardini2024ademamix}, use overlapping exponential averages with hyperparameters controlling their decay rates and contributions.

\paragraph{Learned update parameterization.}
This work approaches learned optimization by predicting gradient-history coefficients through a network shared across parameter groups.
In contrast, typical learned optimizers predict updates for individual coordinates, including the 2016 LSTM optimizer~\citep{andrychowicz2016learning}, RNNprop~\citep{lv2017learning}, VeLO~\citep{metz2022velo}, Celo2~\citep{moudgil2026celo}, and the architectures studied in $\mu$LO~\citep{therien2024mu}.
\citet{li2017learningneural} share LSTM weights within each parameter group but still predict updates for individual coordinates.

\paragraph{Grouped parameter optimization.}
The presented optimizer predicts how to recombine the gradient history of each parameter group.
In contrast, LARS~\citep{you2017large} and LAMB~\citep{you2019large} apply layer-level scaling based on norms, while Adam-mini~\citep{zhang2024adammini} shares second-moment estimates within parameter blocks to determine their adaptive learning rates.

%% file: sections/3_method_policy.tex
\section{Learned Optimization by Dynamic Recombination of Gradient History}
\label{sec:method}

\subsection{Dynamic recombination of gradient history}

\begin{figure}[h]
    \centering
    \includegraphics[width=0.95\linewidth]{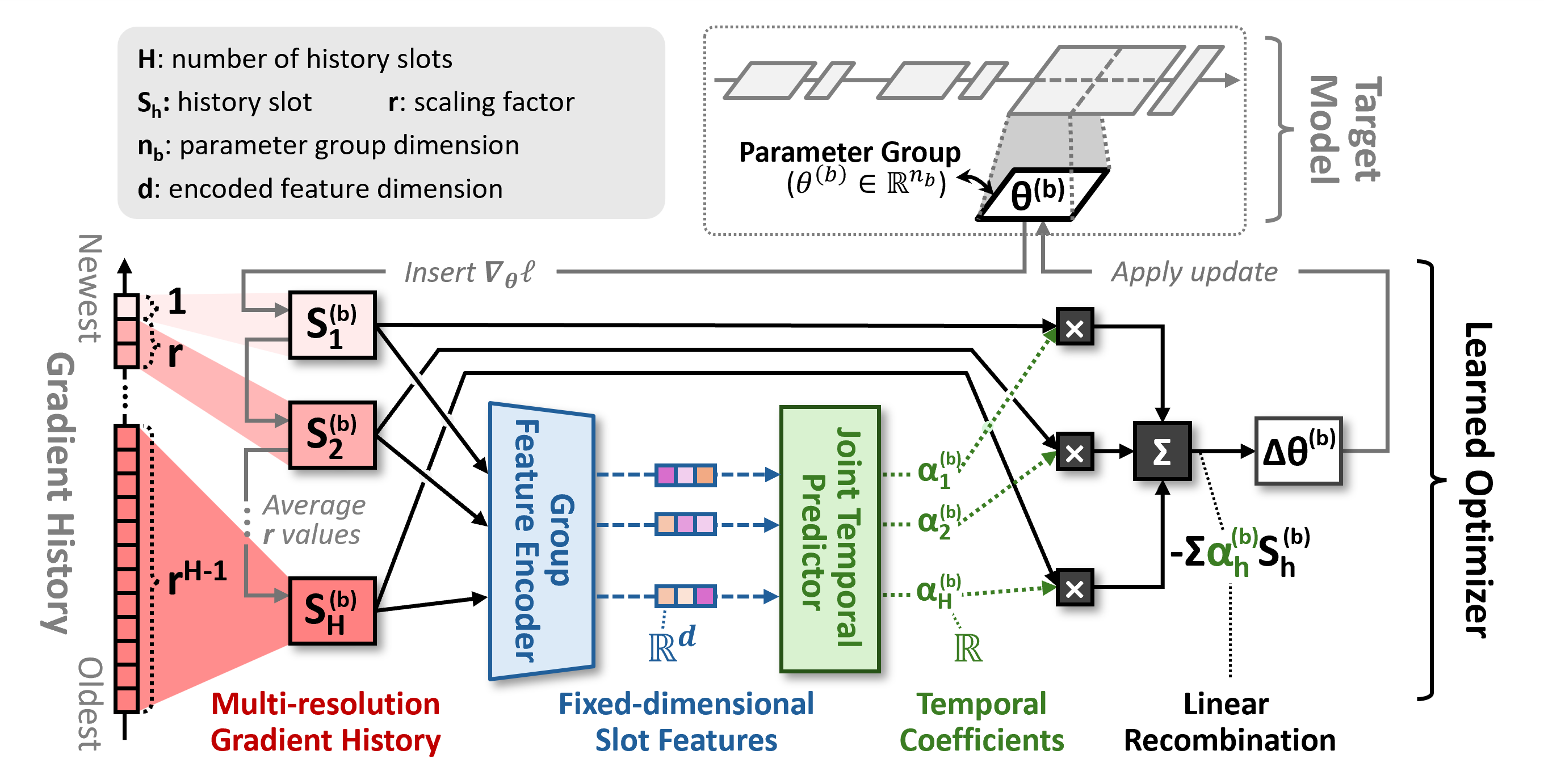}
    \caption{\textbf{Dynamic recombination of multi-resolution gradient history.}
    For each parameter group, the current gradient enters a multi-resolution history.
    A group feature encoder converts each stored average into a fixed-dimensional slot feature, and a
    joint temporal predictor assigns the coefficients used to recombine the history into the group update. Step index $t$ is omitted in the figure as it represents a single optimization step.
    }
    \label{fig:overview}
\end{figure}

The learned optimizer recombines averages of gradient histories from the target model to construct each parameter update, predicting a coefficient for each.
Target-model parameters are partitioned into disjoint groups, where coordinates within each group share the same temporal coefficient vector.

Figure~\ref{fig:overview} illustrates this dataflow from gradient-history storage to parameter update.
At optimization step $t$, parameter group $b$ in the target model denoted as $\smash[b]{\theta_t^{(b)}}$ contains $\smash[b]{n_b}$ parameters , and the target-model loss $\smash[b]{\ell_t}$ yields the gradient $\smash[b]{g_t^{(b)}=\nabla_{\theta_t^{(b)}}\ell_t}$.
Each group maintains \(H\) history slots for disjoint time spans. \(S_{t,h}^{(b)}\) denotes the gradient average stored in slot \(h\), and \(\alpha_{t,h}^{(b)}\) its scalar coefficient.
Let $\smash[b]{\mathcal{A}_t^{(b)}}$ denote the set of populated slots.
The group update is

\begin{equation}
    \Delta\theta_t^{(b)}
    =-\sum_{h\in\mathcal{A}_t^{(b)}}
      \alpha_{t,h}^{(b)}S_{t,h}^{(b)},
    \qquad
    \theta_{t+1}^{(b)}=\theta_t^{(b)}+\Delta\theta_t^{(b)}.
    \label{eq:group-update}
\end{equation}

\subsection{Multi-resolution gradient history}

Multi-resolution gradient history expands the temporal span covered by a small number of slots. 
Each history slot is assigned a tier ($c \in \{1,\dots,C\}$) and can average up to $r^{c-1}$ gradients, where $r\geq2$ is the scaling factor.
As the stored average ages, it is carried forward and averaged into subsequent tiers.

\begin{figure}[h]
    \centering
    \includegraphics[width=\textwidth]{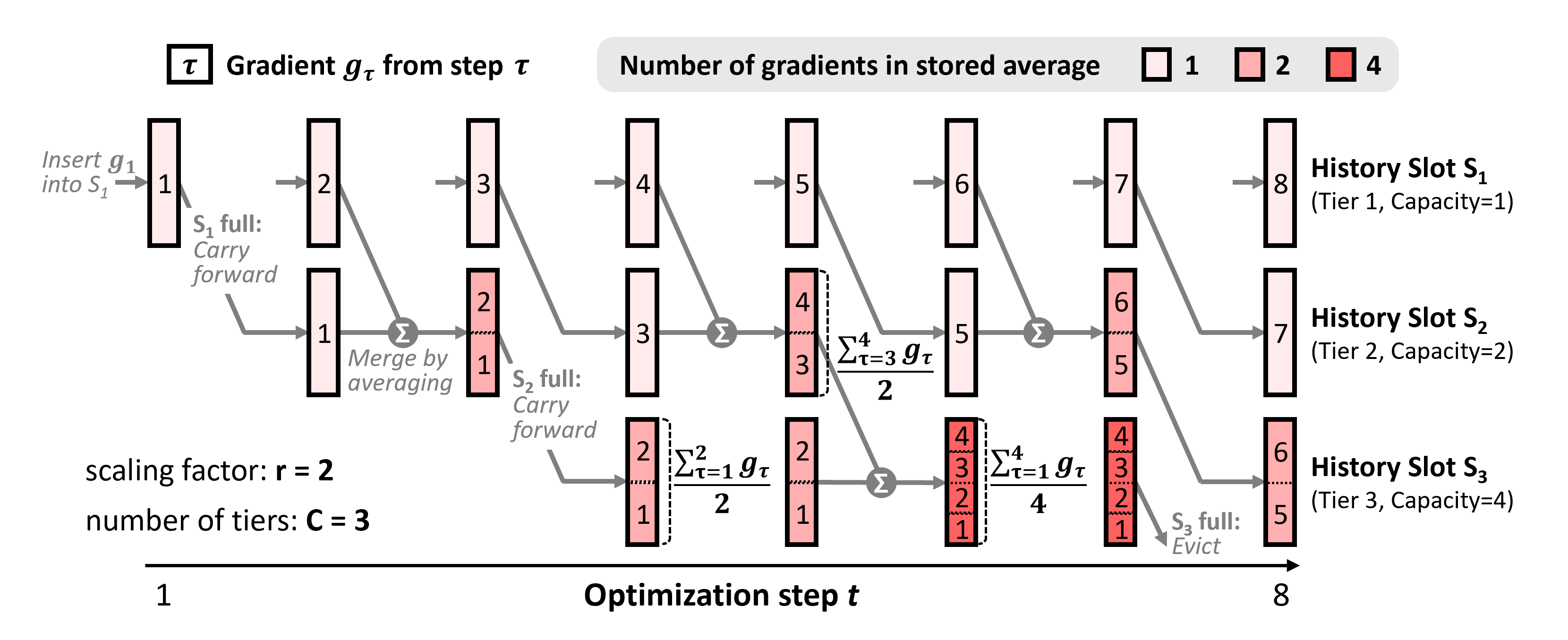}
    \caption{Multi-resolution history with exclusive temporal ranges, tier capacities $1,2,4$, and
    scaling factor $r=2$. At step~8, carry propagates through full slots $S_2$ and $S_3$, evicting
    the previous average over $g_1,\ldots,g_4$ beyond the final tier.}
    \label{fig:multi-resolution-history}
\end{figure}

Figure~\ref{fig:multi-resolution-history} shows an example with $r=2$ and one slot per tier ($c=h$). 
At each step, the current group gradient enters slot $1$, and its displaced value is carried to slot $2$.
In subsequent slots, the carried value either fills an empty slot, averages into a partially filled slot, or replaces a full slot while forwarding its previous average.
A carry beyond slot $H$ is evicted.
Appendix~\ref{app:history-algorithm} gives the complete update procedure.

\subsection{Temporal coefficient prediction}
\label{sec:temporal-coefficient-prediction}

\begin{figure}[h]
    \centering
    \includegraphics[width=\textwidth]{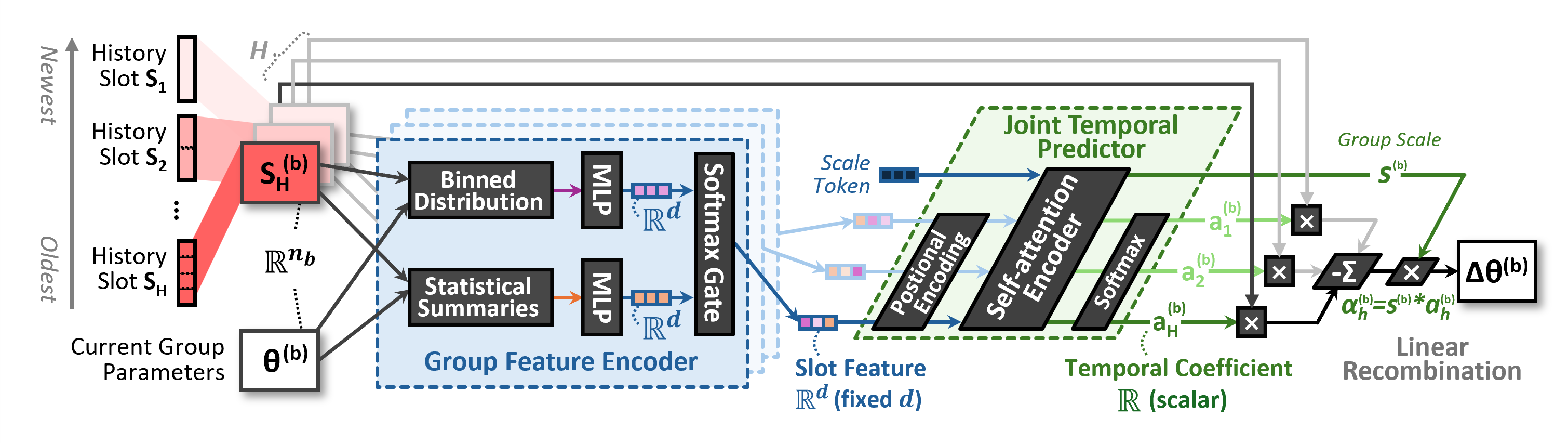}
    \caption{\textbf{Temporal coefficient prediction.} 
    For each history slot, the stored gradient average and current group parameters are converted into histogram features and scalar summaries, then fused into a fixed-dimensional slot feature.
    Positional encoding identifies the represented range, and self-attention processes the slot features jointly.
    Softmax produces normalized temporal allocations \(a_{t,h}^{(b)}\), while a group scale converts them into the final coefficients \(\alpha_{t,h}^{(b)}\) used in Equation~\ref{eq:group-update}.
    Step index $t$ is omitted in the figure as it represents a single optimization step.
    }
    \label{fig:group-representation}
\end{figure}

The temporal coefficients are jointly predicted from each group's current optimization state and stored gradient history.
Figure~\ref{fig:group-representation} depicts the architecture.

\paragraph{Group feature encoder} 
For each parameter group, the group feature encoder summarizes each history slot and the current parameters into a fixed-dimensional slot feature, using their histograms and other statistics such as group dimension, scale, concentration, and alignment.
Separate MLPs project both types of information into \(\mathbb R^d\), where \(d\) is independent of \(n_b\), and a learned two-way Softmax gate fuses them into one slot feature.
Appendix~\ref{app:slot-features} provides its architecture and numerical safeguards.

\paragraph{Joint temporal predictor}
The joint temporal predictor generates the coefficients by processing all slot features via self-attention.
Positional encoding provides tier context, including the center and length of each slot’s history span.
Following self-attention, a coefficient head and Softmax produce nonnegative relative contributions \(a_{t,h}^{(b)}\) that sum to one over \(\mathcal A_t^{(b)}\).
To control the overall update magnitude separately, a scale token uses the history context to predict a group scale \(s_t^{(b)}\).
The final temporal coefficients are \(\alpha_{t,h}^{(b)}=s_t^{(b)}a_{t,h}^{(b)}\), as used in Equation~\ref{eq:group-update}.
Appendix~\ref{app:implementation} specifies the evaluated attention configuration.

\subsection{Trajectory-level optimizer training}

The optimizer is trained to minimize held-out loss throughout an unrolled trajectory of $K$ updates.
Loss is normalized by each meta-task's held-out loss before the unroll, treated as fixed within that unroll, so that tasks with different loss scales contribute comparably.
Let $L_t^{\mathrm{held}}$ be the held-out loss before the unroll and $L_{t+k}^{\mathrm{held}}$ the loss after update $k$.
The meta-loss is

\vspace{-2pt}
\begin{equation}
    \mathcal{L}_{\mathrm{meta}}
    =\sum_{k=1}^{K}
      \frac{L_{t+k}^{\mathrm{held}}}
      {L_t^{\mathrm{held}}}.
    \label{eq:meta-objective}
\end{equation}

Appendix~\ref{app:meta-objective} specifies the numerical safeguards and exact implementation.

%% file: sections/4_evaluation.tex
\section{Evaluation}
\label{sec:evaluation}

\subsection{Experimental Setup}

\paragraph{Optimizer training and deployment.}
The $37{,}389$-parameter learned optimizer is meta-trained on four small network-dataset pairs, which are MLP on synthetic data, ConvNet and a single-block vision transformer on CIFAR-10, and a residual ConvNet on Fashion-MNIST.
The networks contain \(2{,}142\)--\(8{,}450\) parameters.
Six history slots (63-step span) were used during optimizer training, where each meta-epoch consists 160 model-training steps. 
The learned optimizer is evaluated zero-shot on BERT, GPT, Vision Transformer, and GNN architectures unseen during meta-training.
At deployment, the learned optimizer uses history slots with scaling factor $r=2$ to cover the entire target trajectory, referred to as \textit{full-span} unless otherwise stated.
Target network tensors are partitioned into groups of at most 256 coordinates, prioritizing partitions along the output-channel dimension. 

\begin{table*}[t]
    \centering
    \small
    \caption{Comparison with learned-optimizer baselines and fixed-reference Adam on tasks excluded from optimizer training. Language models report terminal validation loss ($\downarrow$). ViT and GNNs report test accuracy at the checkpoint with the lowest validation loss (\%, $\uparrow$). Task results are mean $\pm$ sample standard deviation over three target seeds. Bold and underlined values denote the best and second-best result per row. The aggregate equally weights all 12 tasks, inverting ratios for loss.}
    \label{tab:headline-results}
    \setlength{\tabcolsep}{1.5pt}
    \renewcommand{\arraystretch}{1.40}
    \begin{tabular*}{\textwidth}{@{\extracolsep{\fill}}lrrrrrr}
        \hline
        \multicolumn{7}{l}{\textit{Language models: terminal validation loss} $\downarrow$} \\
        Target & \shortstack{Learned optimizer\\(embeddings, output\\head to Adam)} & \shortstack{Learned optimizer\\(all parameters)} & \shortstack{VeLO\\\scriptsize\citeauthor{metz2022velo}\\\scriptsize\citeyearpar{metz2022velo}} & \shortstack{Celo2\\\scriptsize\citeauthor{moudgil2026celo}\\\scriptsize\citeyearpar{moudgil2026celo}} & \shortstack{$\mu$LO\\\scriptsize\citeauthor{therien2024mu}\\\scriptsize\citeyearpar{therien2024mu}} & Adam \\
        \hline
        BERT-Tiny & \textbf{3.462 $\pm$ .058} & \underline{3.755 $\pm$ .187} & 3.846 $\pm$ .034 & 5.162 $\pm$ .209 & 3.808 $\pm$ .246 & 3.781 $\pm$ .033 \\
        GPT-Tiny & \underline{4.280 $\pm$ .032} & 4.330 $\pm$ .027 & \textbf{4.229 $\pm$ .028} & 4.540 $\pm$ .028 & 4.543 $\pm$ .032 & 4.285 $\pm$ .026 \\
        \hline
    \end{tabular*}
    \vspace{4pt}
    \begin{tabular*}{\textwidth}{@{\extracolsep{\fill}}llrrrrr}
        \hline
        \multicolumn{7}{l}{\textit{Vision and graph models: validation-selected test accuracy (\%)} $\uparrow$} \\
        Data & Model & Learned optimizer & \shortstack{VeLO\\\scriptsize\citeauthor{metz2022velo}\\\scriptsize\citeyearpar{metz2022velo}} & \shortstack{Celo2\\\scriptsize\citeauthor{moudgil2026celo}\\\scriptsize\citeyearpar{moudgil2026celo}} & \shortstack{$\mu$LO\\\scriptsize\citeauthor{therien2024mu}\\\scriptsize\citeyearpar{therien2024mu}} & Adam \\
        \hline
        CIFAR-100 & ViT & \underline{31.15 $\pm$ .23} & \textbf{35.20 $\pm$ 1.38} & 15.88 $\pm$ .22 & 20.46 $\pm$ .37 & 25.36 $\pm$ 3.14 \\
        \hline
        Cora & GCN & \textbf{81.70 $\pm$ .20} & 80.47 $\pm$ .58 & 80.17 $\pm$ .67 & \underline{80.80 $\pm$ .36} & 79.27 $\pm$ .83 \\
             & GraphSAGE & 79.60 $\pm$ .44 & \underline{79.93 $\pm$ 1.01} & 79.20 $\pm$ .20 & \textbf{80.17 $\pm$ .32} & 78.73 $\pm$ .32 \\
             & GAT & \underline{81.97 $\pm$ .31} & 80.77 $\pm$ 1.19 & 80.27 $\pm$ 1.46 & \textbf{82.40 $\pm$ .72} & 78.27 $\pm$ 1.26 \\
        \hline
        CiteSeer & GCN & \underline{71.03 $\pm$ .81} & 68.53 $\pm$ .55 & 68.43 $\pm$ .78 & \textbf{71.23 $\pm$ .83} & 66.20 $\pm$ 1.61 \\
                 & GraphSAGE & \underline{69.87 $\pm$ .87} & 69.23 $\pm$ 1.01 & 68.03 $\pm$ .21 & \textbf{70.00 $\pm$ .36} & 65.57 $\pm$ 1.36 \\
                 & GAT & \textbf{69.00 $\pm$ .26} & 68.07 $\pm$ .38 & 68.17 $\pm$ .55 & \underline{68.77 $\pm$ .80} & 66.20 $\pm$ .82 \\
        \hline
        PubMed & GCN & \textbf{78.67 $\pm$ .21} & 77.83 $\pm$ .21 & 75.80 $\pm$ .35 & \underline{78.53 $\pm$ .25} & 77.50 $\pm$ .36 \\
               & GraphSAGE & \underline{77.37 $\pm$ .49} & 75.93 $\pm$ .67 & 73.40 $\pm$ .26 & \textbf{77.40 $\pm$ .26} & 75.17 $\pm$ .35 \\
               & GAT & \textbf{78.73 $\pm$ .31} & 77.63 $\pm$ .55 & 76.90 $\pm$ .95 & \underline{78.53 $\pm$ .31} & 77.10 $\pm$ 1.42 \\
        \noalign{\hrule height 1.1pt}
        \multicolumn{2}{l}{Arithmetic mean across 9 cells} & \textbf{76.44 $\pm$ .28} & 75.38 $\pm$ .43 & 74.49 $\pm$ .36 & \underline{76.43 $\pm$ .14} & 73.78 $\pm$ .26 \\
        \hline
    \end{tabular*}
    \vspace{3pt}
    \begin{tabular*}{\textwidth}{@{\extracolsep{\fill}}lccccc}
        \hline
        Across 12 tasks & Learned optimizer$^*$ & \shortstack{VeLO\\\scriptsize\citeauthor{metz2022velo}\\\scriptsize\citeyearpar{metz2022velo}} & \shortstack{Celo2\\\scriptsize\citeauthor{moudgil2026celo}\\\scriptsize\citeyearpar{moudgil2026celo}} & \shortstack{$\mu$LO\\\scriptsize\citeauthor{therien2024mu}\\\scriptsize\citeyearpar{therien2024mu}} & Adam \\
        \hline
        \noalign{\vskip 3pt}
        \shortstack[l]{Geometric-mean gain over\\fixed-reference Adam (\%) $\uparrow$} & \textbf{+5.32} & \underline{+4.46} & $-6.01$ & +0.38 & \textendash \\[3pt]
        \hline
    \end{tabular*}
    \par\vspace{2pt}
    {\footnotesize\raggedright $^*$Embeddings and the output head are trained with Adam for language models.\par}
\end{table*}

\begin{table*}[t]
    \centering
    \small
    \caption{Comparison with task-tuned classical optimizers. Language models report terminal validation loss ($\downarrow$). ViT and GNNs report test accuracy at the checkpoint with the lowest validation loss (\%, $\uparrow$). Baseline hyperparameters are selected by validation loss and then held fixed across evaluation seeds. Task results are mean $\pm$ sample standard deviation over three target seeds. Bold and underlined values denote the best and second-best result per row. The aggregate equally weights all 12 tasks, inverting ratios for loss.}
    \label{tab:additional-optimizer-comparison}
    \setlength{\tabcolsep}{1.5pt}
    \renewcommand{\arraystretch}{1.40}
    \begin{tabular*}{\textwidth}{@{\extracolsep{\fill}}lrrrrrr}
        \hline
        \multicolumn{7}{l}{\textit{Language models: terminal validation loss} $\downarrow$} \\
        Target & \shortstack{Learned optimizer\\(embeddings, output\\head to Adam)} & \shortstack{Learned optimizer\\(all parameters)} & \shortstack{Task-tuned\\Adam} & \shortstack{Task-tuned\\AdamW} & \shortstack{Task-tuned\\SGDM} & \shortstack{Task-tuned\\AdEMAMix\\\scriptsize\citeauthor{pagliardini2024ademamix}\\\scriptsize\citeyearpar{pagliardini2024ademamix}} \\
        \hline
        BERT-Tiny & \textbf{3.462 $\pm$ .058} & 3.755 $\pm$ .187 & \underline{3.636 $\pm$ .194} & 3.778 $\pm$ .033 & 4.118 $\pm$ .054 & 3.773 $\pm$ .010 \\
        GPT-Tiny & 4.280 $\pm$ .032 & 4.330 $\pm$ .027 & \underline{4.270 $\pm$ .029} & 4.306 $\pm$ .035 & 4.442 $\pm$ .024 & \textbf{4.246 $\pm$ .028} \\
        \hline
    \end{tabular*}
    \vspace{4pt}
    \begin{tabular*}{\textwidth}{@{\extracolsep{\fill}}llrrrrr}
        \hline
        \multicolumn{7}{l}{\textit{Vision and graph models: validation-selected test accuracy (\%)} $\uparrow$} \\
        Data & Model & Learned optimizer & \shortstack{Task-tuned\\Adam} & \shortstack{Task-tuned\\AdamW} & \shortstack{Task-tuned\\SGDM} & \shortstack{Task-tuned\\AdEMAMix\\\scriptsize\citeauthor{pagliardini2024ademamix}\\\scriptsize\citeyearpar{pagliardini2024ademamix}} \\
        \hline
        CIFAR-100 & ViT & 31.15 $\pm$ .23 & 32.40 $\pm$ 1.27 & \textbf{33.49 $\pm$ 1.40} & \underline{32.71 $\pm$ .33} & 30.96 $\pm$ .33 \\
        \hline
        \input{figures/data/TableB_GNNTaskTunedRows.tex}
    \end{tabular*}
    \vspace{6pt}
    \begin{tabular*}{\textwidth}{@{\extracolsep{\fill}}lccccc}
        \hline
        Across 12 tasks & \shortstack{Learned\\optimizer$^*$} & \shortstack{Task-tuned\\Adam} & \shortstack{Task-tuned\\AdamW} & \shortstack{Task-tuned\\SGDM} & \shortstack{Task-tuned\\AdEMAMix\\\scriptsize\citeauthor{pagliardini2024ademamix}\\\scriptsize\citeyearpar{pagliardini2024ademamix}} \\
        \hline
        \noalign{\vskip 3pt}
        \shortstack[l]{Geometric-mean gain over\\fixed-reference Adam (\%) $\uparrow$} & \textbf{+5.32} & +2.91 & +2.84 & \underline{+3.72} & +2.18 \\[3pt]
        \hline
    \end{tabular*}
    \par\vspace{2pt}
    {\footnotesize\raggedright $^*$Embeddings and the output head are trained with Adam for language models.\par}
\end{table*}

\paragraph{Baselines and pairing.}
The fixed Adam reference follows the recommended hyperparameters from its paper, where learning rate is $10^{-3}$, $(\beta_1,\beta_2)=(0.9,0.999)$, and $\epsilon=10^{-8}$ \citep{kingma2014adam}.
It uses zero weight decay and no learning-rate schedule.
Table~\ref{tab:headline-results} compares our optimizer with VeLO, Celo2, $\mu$LO, and fixed-reference Adam without target-specific optimizer training; Celo2 follows the recommended deployment configuration from its authors.
Celo2 assigns one-dimensional tensors to AdamW, as well as embeddings and output heads on language targets.
$\mu$LO uses released optimizer weights and maximal update parameterization ($\mu$P) initialization and scaling, updating all target parameters without task-specific tuning~\citep{therien2024mu}.
Table~\ref{tab:additional-optimizer-comparison} compares against Adam, AdamW, momentum stochastic gradient descent (SGDM), and AdEMAMix after selecting their hyperparameters by validation loss and holding each selection fixed across evaluation seeds.
Where applicable, paired comparisons share the same initialization, data ordering, stochastic streams, number of optimizer updates, and parameter grouping.

\paragraph{Target tasks.}
BERT-Tiny ($4.42$M parameters) follows \citet{turc2019wellread}, while GPT-Tiny
($6.85$M parameters) uses a GPT-2 decoder~\citep{radford2019language}.
Both have two Transformer layers of width 128 and two attention heads.
BERT-Tiny is trained for masked-token prediction on Wikipedia--BookCorpus, and GPT-Tiny for next-token prediction on WikiText-103.
Further evaluations extend both Tiny models to 250k updates and increase the target model size to BERT-Base ($109.5$M parameters) over 100k updates.
The vision target expands the single-block optimizer-training model to a $4.81$M-parameter,
six-block, width-256, eight-head transformer on CIFAR-100.
The graph evaluation crosses GCN, GraphSAGE, and GAT with Cora, CiteSeer, and
PubMed, yielding nine node-classification models with $32$k--$475$k parameters~\citep{kipf2016semi,hamilton2017inductive,velickovic2018graph,sen2008collective}.
Test accuracies for both ViT and GNN tasks were reported on the checkpoint with the lowest validation loss. 
Appendix~\ref{app:implementation} specifies the complete architectures and training protocols.

\subsection{Zero-Shot Generalization to Unseen Tasks}

\paragraph{Language models.}
The learned optimizer generalizes from 160-step classification tasks to unseen language objectives, datasets, and architectures without retraining.
On BERT-Tiny, it outperforms Adam and the compared learned optimizers even when applied to all parameters.
Assigning embeddings and the output head to Adam further improves performance and gives lower validation loss than the task-tuned baselines (Tables~\ref{tab:headline-results} and~\ref{tab:additional-optimizer-comparison}).

This generalization extends to longer training horizons and larger models.
At 250k updates, applying the learned optimizer to all parameters retains its advantage over Adam on BERT-Tiny and achieves comparable performance on GPT-Tiny.
On BERT-Base, it matches Adam without optimizer retraining when embeddings and the output head are assigned to Adam.
The learned optimizer updates 85.1M parameters, approximately 10,000 times the parameter count of the largest model used during meta-training (Table~\ref{tab:extended-language}).
On GPT-Tiny, the learned optimizer with embeddings and the output head trained with Adam also achieves lower validation loss than VeLO at the same training-time budget corresponding to its 250k-update completion (Appendix~\ref{app:extended-language}, Figure~\ref{fig:gpt-250k-time-matched}).

\begin{table*}[t]
    \centering
    \small
    \caption{Generalization to longer trajectories and a larger language model. Terminal validation
    loss ($\downarrow$) is reported as mean $\pm$ sample standard deviation over three seeds.
    Both parameter assignments use the same Adam reference for each target.
    Bold marks the lowest mean, including ties at the displayed precision.
    Appendix~\ref{app:extended-language} gives the training protocols.}
    \label{tab:extended-language}
    \setlength{\tabcolsep}{5pt}
    \renewcommand{\arraystretch}{1.40}
    \begin{tabular*}{\textwidth}{@{\extracolsep{\fill}}llrrr}
        \hline
        Target & Learned optimizer & Updates & Learned & Adam \\
        \hline
        \input{figures/data/TableB_ExtendedLanguageRows.tex}
    \end{tabular*}
\end{table*}

\paragraph{Graph models.}
\label{sec:gnn-evaluation}

The learned optimizer generalizes from minibatch classification to full-batch training on unseen graph architectures and datasets.
It achieves the highest test accuracy in four of nine comparisons with learned optimizers and the highest average accuracy against task-tuned classical optimizers (Tables~\ref{tab:headline-results} and~\ref{tab:additional-optimizer-comparison}).

\paragraph{Vision transformer.}
The learned optimizer generalizes from a single-block model on CIFAR-10 to a six-block ViT on CIFAR-100 without retraining.
It achieves higher test accuracy than Adam, extending generalization to both a larger vision model and an unseen dataset (Table~\ref{tab:headline-results}).
AdamW achieves the highest accuracy among the task-tuned classical optimizers, but its 17-run hyperparameter search required 7.52 GPU-hours of training updates, 8.6 times the 0.87 GPU-hours used to meta-train the presented optimizer (Table~\ref{tab:additional-optimizer-comparison}).

\begin{figure}[!t]
    \centering
    \includegraphics[width=\textwidth]{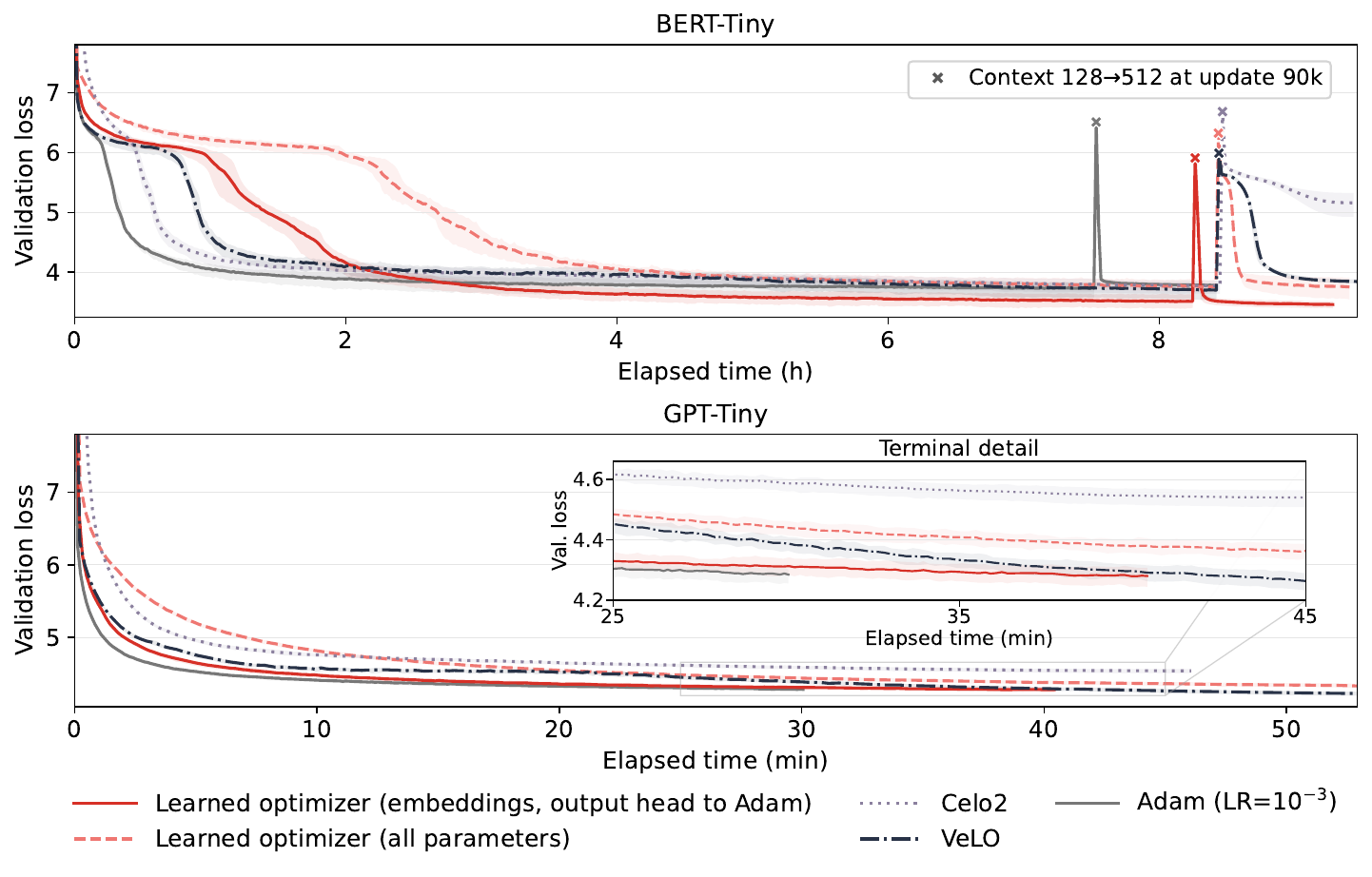}
    \caption{Language-model validation loss against training time calculated from measured step times. BERT-Tiny uses its
    measured context-128 step time through update 90,000 and context-512 step time
    thereafter; crosses mark the transition on every curve. GPT-Tiny extends through
    VeLO's terminal time, and its inset magnifies 25--45 minutes. The GPT curve for the learned optimizer applied to all parameters continues beyond the displayed range. Lines show three-seed means, and bands
    show pointwise minimum--maximum ranges. Solid red applies the learned optimizer while
    Adam updates embedding and output-head tensors using the fixed-reference Adam
    hyperparameters; dashed red applies the learned optimizer to all parameters.}
    \label{fig:time-matched-language}
\end{figure}

\subsection{Time-Matched Performance and Deployment Cost}

\paragraph{Time-matched trajectories.}
With embeddings and the output head assigned to Adam, it completes BERT-Tiny training with lower validation loss in less time than Celo2 and VeLO (Figure~\ref{fig:time-matched-language}).
On GPT-Tiny, it also reaches lower validation loss than both at its completion time, although VeLO achieves lower loss with additional training.
Appendix Figure~\ref{fig:gnn-time-matched-complete} reports the comparisons for all three Cora graph architectures.

\begin{figure}[!p]
    \centering
    \begin{subfigure}[t]{0.485\textwidth}
        \centering
        \includegraphics[width=0.80\linewidth]{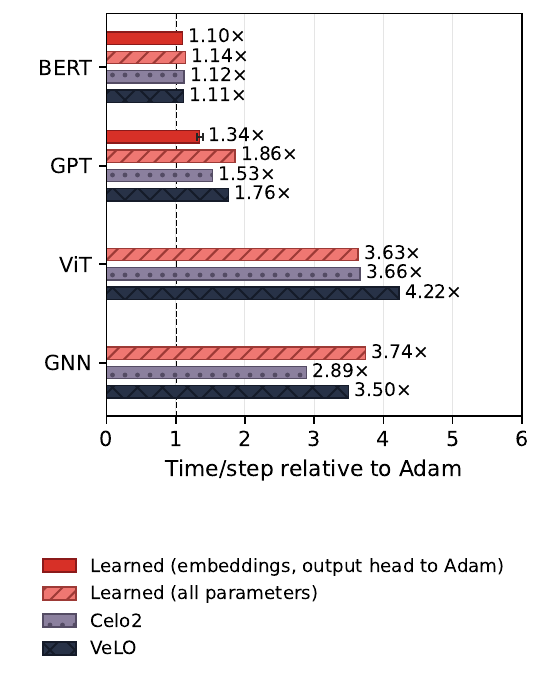}
        \caption{Time/step relative to Adam.}
        \label{fig:deployment-time}
    \end{subfigure}\hfill
    \begin{subfigure}[t]{0.485\textwidth}
        \centering
        \includegraphics[width=0.80\linewidth]{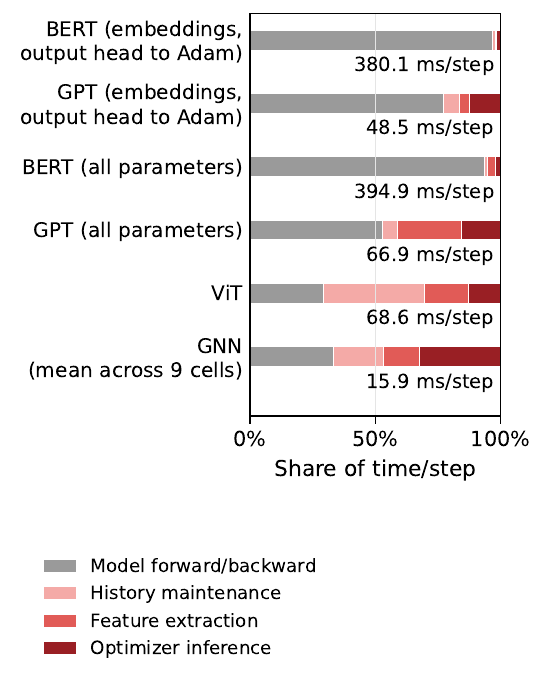}
        \caption{Time/step breakdown for the learned optimizer.}
        \label{fig:deployment-time-breakdown}
    \end{subfigure}
    \caption{Deployment time on an NVIDIA B200.
    \textbf{(a)} Celo2 uses its recommended mixed parameter routing; the GPT whisker
    spans two profiles. \textbf{(b)} Labels give measured total time/step. History
    maintenance includes updating stored gradients and assembling parameter groups. Feature extraction
    includes history normalization and gradient statistics. Optimizer inference includes
    optimizer inference, history recombination, and parameter application. Across the nine GNN tasks,
    panel (a) uses geometric means of time ratios and panel (b) uses arithmetic means of time per step.
    Appendix~\ref{app:resource-measurements} defines the profiles.}
    \label{fig:deployment-cost}
    \vspace{3pt}
    \includegraphics[width=0.98\linewidth]{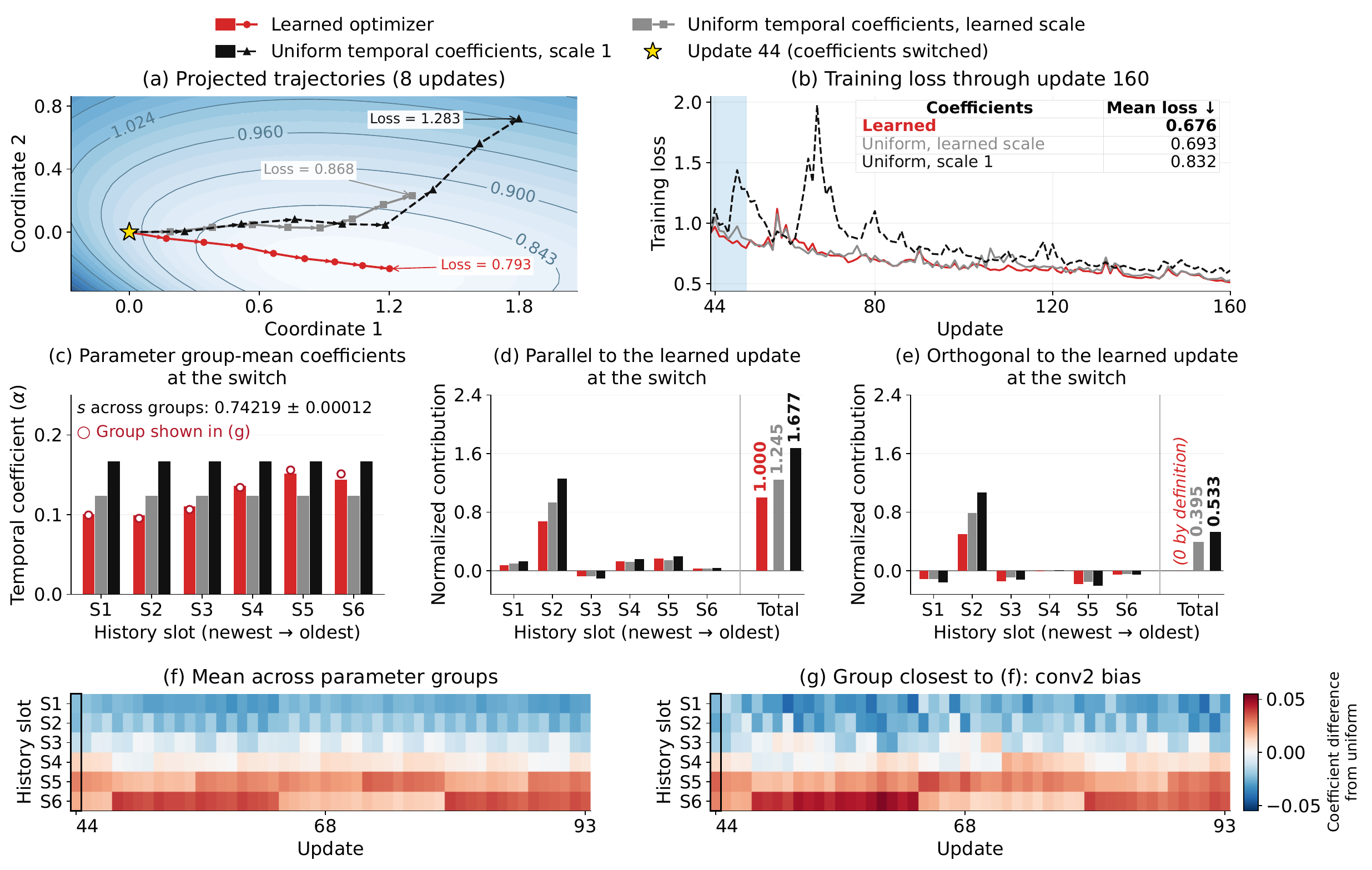}
    \caption{\textbf{History recombination changes optimization behavior.}
    Temporal coefficients are switched across all LeNet groups before update 44.
    (a) Projected trajectories, with actual training loss at the endpoints.
    (b) Full-training-set loss through update 160, with (a)'s window shaded. Inset: normalized loss AUC over updates 43--160.
    (c) Parameter group-mean temporal coefficients $\alpha$ at the switch. Circles denote (g), and $s$ reports mean $\pm$ SD.
    (d--e) Parallel and orthogonal slot contributions at the switch, normalized by the learned full-network update norm.
    (f--g) Differences from same-scale uniform temporal coefficients: group mean and the group closest to (f) over updates 44--93.
    Appendix Figure~\ref{fig:group-coefficient-variation} defines selection.}
    \label{fig:uniform-coefficient-branches}

\end{figure}

\paragraph{Optimizer training and deployment time.}
Training the learned optimizer requires 0.87 B200 GPU-hours, and the same optimizer is deployed across all target tasks without retraining.
During deployment, history management and feature extraction grow with parameter count, while self attention and coefficient and scale prediction depend on the number of parameter groups and history slots.
Figure~\ref{fig:deployment-cost} reports the deployment time and separates model computation from these optimizer operations.

\paragraph{History slot scaling.}
History storage can be limited without restricting deployment to short training horizons.
Increasing the scale factor $r$ averages more gradients per slot, extending history coverage while keeping the number of stored gradient averages unchanged.
Six slots retain competitive performance across the evaluated tasks (Table~\ref{tab:temporal-range}).

\begin{table}[H]
    \captionsetup{font=small}
    \centering
    \small
    \caption{Effect of history span. Language models report terminal validation loss ($\downarrow$).
    ViT and GNNs report test accuracy (\%, $\uparrow$) at the checkpoint with the lowest validation loss.
    Values are mean $\pm$ sample standard deviation across three target seeds.
    The GNN row averages the nine model--dataset comparisons within each seed.
    Bold and underlined values mark the best and second-best result in each row. Both GNN history spans are bold because their mean $\pm$ standard deviation intervals overlap.}
    \label{tab:temporal-range}
    \setlength{\tabcolsep}{3pt}
    \renewcommand{\arraystretch}{1.20}
    \begin{tabular*}{\textwidth}{@{\extracolsep{\fill}}llrrr}
        \hline
        Target & Learned optimizer & Full span & 63-step span & Adam \\
        \hline
        BERT-Tiny & \shortstack[l]{Embeddings, output\\head to Adam} & \underline{3.462 $\pm$ 0.058} & \textbf{3.456 $\pm$ 0.061} & 3.781 $\pm$ 0.033 \\
        BERT-Tiny & All parameters & \underline{3.755 $\pm$ 0.187} & \textbf{3.750 $\pm$ 0.166} & 3.781 $\pm$ 0.033 \\
        GPT-Tiny & \shortstack[l]{Embeddings, output\\head to Adam} & \textbf{4.280 $\pm$ 0.032} & 4.295 $\pm$ 0.030 & \underline{4.285 $\pm$ 0.026} \\
        GPT-Tiny & All parameters & \underline{4.330 $\pm$ 0.027} & 4.363 $\pm$ 0.026 & \textbf{4.285 $\pm$ 0.026} \\
        \hline
        ViT/CIFAR-100 & All parameters & \underline{31.15 $\pm$ 0.23} & \textbf{31.36 $\pm$ 0.48} & 25.36 $\pm$ 3.14 \\
        \hline
        \rule{0pt}{5ex}\shortstack[l]{GNN arithmetic mean\\across 9 cells}
                      & All parameters & \textbf{76.44 $\pm$ 0.28} & \textbf{76.47 $\pm$ 0.24} & 73.78 $\pm$ 0.26 \\
        \hline
    \end{tabular*}
\end{table}

\clearpage
\subsection{Ablations and Behavioral Analysis}
\label{sec:history-controls}

\paragraph{Behavioral analysis of temporal recombination.}
Switching from learned temporal coefficients $\alpha$ to uniform temporal coefficients changes the optimization trajectory and its subsequent training loss (Figure~\ref{fig:uniform-coefficient-branches}a--b).
In this LeNet example, both uniform-coefficient continuations show loss spikes and higher mean training loss, whether retaining the learned scale or fixing it to 1.
At the coefficient switch, differences in temporal coefficients change both update magnitude and direction even at the same scale (c--e).
Temporal coefficients evolve over time in both the group mean and the group closest to that mean (f--g), rather than following a static weighting pattern.
Appendix Figure~\ref{fig:group-coefficient-variation} shows how these temporal patterns vary across parameter groups.

\paragraph{Temporal recombination and architectural ablations.}
Learned recombination can contribute to training stability.
On LeNet/Fashion-MNIST, switching from learned coefficients to uniform coefficients after updates 5 or 15, while retaining the learned scale, increases training-loss AUC through update 200 by 16.4\% and 17.0\%, respectively.
With uniform coefficients and the scale also fixed to 1 at the switch, AUC increases by 37.0\% and 71.5\% relative to the unchanged learned optimizer (Appendix Figure~\ref{fig:uniform-coefficient-branches-both} and Table~\ref{tab:coefficient-branch-auc}).
On ViT, with the scale fixed to 1 for both policies, all three learned-coefficient runs complete 50 epochs without numerical divergence, whereas all three uniform-coefficient runs diverge with nonfinite gradients.
Separately, retraining without self-attention reduces mean GAT/PubMed test accuracy from 78.77\% to 49.00\% across three target seeds.
Appendix~\ref{app:coefficient-scale-controls} reports these ablations.

\paragraph{History span ablations.}
Limiting history to six slots retains competitive performance with Adam while reducing history storage (Table~\ref{tab:temporal-range}).
These results show that the optimizer can remain effective with a restricted history span.
GNN mean accuracy is nearly identical for both spans. Appendix~\ref{app:history-depth} reports all nine GNN comparisons.

%% file: figures/data/TableB_GNNTaskTunedRows.tex
        Cora & GCN & \textbf{81.70 $\pm$ .20} & 79.37 $\pm$ .81 & 80.73 $\pm$ .85 & \underline{81.13 $\pm$ .72} & 79.43 $\pm$ .75 \\
             & GraphSAGE & \textbf{79.60 $\pm$ .44} & 78.80 $\pm$ .17 & 78.23 $\pm$ 1.19 & \textbf{79.60 $\pm$ 1.01} & \underline{78.83 $\pm$ .25} \\
             & GAT & \textbf{81.97 $\pm$ .31} & 80.37 $\pm$ 1.97 & 80.67 $\pm$ 1.06 & \underline{81.57 $\pm$ 1.27} & 80.10 $\pm$ 1.73 \\
        \hline
        CiteSeer & GCN & \textbf{71.03 $\pm$ .81} & 66.93 $\pm$ .32 & 67.53 $\pm$ .65 & \underline{70.93 $\pm$ .71} & 67.00 $\pm$ .36 \\
                 & GraphSAGE & \underline{69.87 $\pm$ .87} & 67.30 $\pm$ 1.14 & 65.93 $\pm$ 1.44 & \textbf{70.13 $\pm$ .45} & 67.40 $\pm$ 1.23 \\
                 & GAT & \underline{69.00 $\pm$ .26} & 66.90 $\pm$ 1.15 & 67.47 $\pm$ 1.18 & \textbf{70.37 $\pm$ .23} & 66.97 $\pm$ 1.52 \\
        \hline
        PubMed & GCN & \textbf{78.67 $\pm$ .21} & 77.23 $\pm$ .23 & 77.20 $\pm$ .30 & \underline{77.73 $\pm$ .67} & 76.97 $\pm$ .40 \\
               & GraphSAGE & \textbf{77.37 $\pm$ .49} & 75.07 $\pm$ .81 & 74.90 $\pm$ .20 & \underline{77.17 $\pm$ .84} & 74.63 $\pm$ .68 \\
               & GAT & \textbf{78.73 $\pm$ .31} & 75.97 $\pm$ .31 & 75.80 $\pm$ .20 & \underline{77.37 $\pm$ .32} & 75.90 $\pm$ .36 \\
        \noalign{\hrule height 1.1pt}
        \multicolumn{2}{l}{Arithmetic mean across 9 cells} & \textbf{76.44 $\pm$ .28} & 74.21 $\pm$ .27 & 74.27 $\pm$ .34 & \underline{76.22 $\pm$ .33} & 74.14 $\pm$ .24 \\
        \hline

%% file: figures/data/TableB_ExtendedLanguageRows.tex
BERT-Tiny & Embeddings, output head to Adam & 250k & \textbf{3.404 $\pm$ 0.057} & 3.736 $\pm$ 0.043 \\
BERT-Tiny & All parameters & 250k & \textbf{3.544 $\pm$ 0.201} & 3.736 $\pm$ 0.043 \\
\hline
GPT-Tiny & Embeddings, output head to Adam & 250k & \textbf{4.185 $\pm$ 0.032} & \textbf{4.185 $\pm$ 0.028} \\
GPT-Tiny & All parameters & 250k & 4.199 $\pm$ 0.027 & \textbf{4.185 $\pm$ 0.028} \\
\hline
BERT-Base & Embeddings, output head to Adam & 100k & \textbf{7.410 $\pm$ 0.006} & \textbf{7.410 $\pm$ 0.006} \\
\hline

%% file: sections/6_conclusion.tex
\section{Conclusion}
\label{sec:conclusion}

This paper demonstrates that a compact temporal recombination policy, meta-trained only on small classification tasks, can generalize to unseen tasks. 
These results establish recombination of gradient history as a practical design principle for learned optimization. 

%% file: sections/A_evaluation_details.tex
\section{Implementation and Evaluation Details}
\label{app:implementation}

This appendix specifies the evaluated optimizer, optimizer-training procedure, target protocols,
and resource measurements.

\subsection{Learned-optimizer configuration}

Each parameter tensor is partitioned into fixed groups of at most 256 active coordinates. For
tensors with at least two dimensions, trailing dimensions are flattened within each leading-dimension
slice. Adjacent complete slices are packed up to the limit, and oversized slices are split into
contiguous segments. One-dimensional tensors are segmented directly. Final incomplete segments are
padded for batched optimizer inference, with padded coordinates masked from features and updates.

Meta-training uses six history slots with capacities $1,2,4,8,16,32$ and $r=2$.
Full-span deployment increases the number of slots to cover the target training horizon.
The six-slot controls retain the meta-training capacities.
Manuscript indices run newest to oldest. Attention receives the reverse order. Each populated slot
is paired with its temporal center and stored-gradient count.

\subsubsection{History-slot feature definitions}
\label{app:slot-features}

For a group of $n_b$ parameters, let $S=S_{t,h}^{(b)}$ denote a populated stored average and
$\theta=\theta_t^{(b)}$ its current parameter vector. Table~\ref{tab:slot-features} defines the six
statistics computed per slot. The stabilized norms are
$\operatorname{RMS}_{\epsilon}(v)=\sqrt{\lVert v\rVert_2^2/n_b+\epsilon}$ and
$\lVert v\rVert_{2,\epsilon}=\sqrt{\lVert v\rVert_2^2+\epsilon}$.

\begin{table}[H]
    \centering
    \small
    \caption{Statistics computed for the stored gradient average in each populated history slot.}
    \label{tab:slot-features}
    \begin{tabularx}{\linewidth}{p{0.29\linewidth}X}
        \hline
        Feature & Definition \\
        \hline
        Group size & $\log_{10}n_b$ \\
        Stored-average scale & $\log_{10}\operatorname{RMS}_{\epsilon}(S)$ \\
        Current-parameter scale & $\log_{10}\operatorname{RMS}_{\epsilon}(\theta)$ \\
        Coordinate uniformity & $\lVert S\rVert_1/(\sqrt{n_b}\lVert S\rVert_{2,\epsilon}+\epsilon)$ \\
        Gradient--parameter alignment & $\langle S,\theta\rangle/(\lVert S\rVert_{2,\epsilon}\lVert\theta\rVert_{2,\epsilon}+\epsilon)$ \\
        Gradient-to-parameter scale & $\log_{10}\operatorname{RMS}_{\epsilon}(S)-\log_{10}\operatorname{RMS}_{\epsilon}(\theta)$ \\
        \hline
    \end{tabularx}
\end{table}

The histogram path normalizes each populated slot by its gradient RMS. For normalized coordinate
$\widetilde S_i$ and current parameter $\theta_i$, the seven tuple entries are
$\operatorname{sign}(\widetilde S_i)\log_{10}(1+|\widetilde S_i|)$,
$\log_{10}(|\widetilde S_i|+\epsilon)$, $\log_{10}(|\theta_i|+\epsilon)$,
$\log_{10}((|\widetilde S_i|+\epsilon)/(|\theta_i|+\epsilon))$, a signed logarithm of
$\widetilde S_i$ relative to the slot RMS, $\operatorname{sign}(\widetilde S_i)\operatorname{sign}(\theta_i)$,
and the normalized group index in the extraction batch. Tuple channels are standardized within each
slot and clipped to $[-5,5]$. Sixteen bins are indexed by
$\log_{10}(|\widetilde S_i|+\epsilon)$ and store their assigned mean tuples. Empty bins are zero.
Padded coordinates are excluded, and feature extraction is detached during optimizer training.

Two-layer MLPs with intermediate LayerNorm and GELU project the six statistics and 112-dimensional
histogram to separate 32-dimensional vectors. LayerNorm precedes their learned two-way gated fusion.
A fixed sinusoidal encoding adds temporal center and stored-gradient count. Unpopulated slots are
excluded from attention and coefficient normalization.

The temporal predictor uses two width-32 Transformer encoder layers, four attention heads, and
feed-forward width 128. Softmax normalizes coefficients over populated slots. The group scale is
incorporated into the coefficients as specified in Section~\ref{sec:temporal-coefficient-prediction}. Its learned
scale vector attends to every populated slot, while an asymmetric mask prevents slot tokens from
attending to it. For its contextualized representation
$h_{t,\mathrm{scale}}^{(b)}\in\mathbb R^d$ and scalar projection
$q:\mathbb R^d\rightarrow\mathbb R$, the scale is
\[
    s_t^{(b)}
    =2\,\sigma\!\left(
      z_0+\epsilon\tanh\!\left(q(h_{t,\mathrm{scale}}^{(b)})\right)
    \right)\in(0,2),
\]
where $z_0$ is learned and $\epsilon=0.5$. The scale is initialized at one and multiplies the
completed recombination.

\subsubsection{Multi-resolution history update}
\label{app:history-algorithm}

Algorithm~\ref{alg:history-update} gives the update for one group, omitting group indices.
$c_{t,h}$ counts values accumulated at tier $h$. Slots and counters start empty or zero, and a carry
beyond tier $H$ is discarded.

\begin{algorithm}[t]
\caption{One-slot-per-tier multi-resolution history update at optimization step $t$.
Processing stops when no value is carried to the next slot.}
\label{alg:history-update}
\begin{algorithmic}[1]
    \REQUIRE $\begin{array}[t]{ll}
        g_t & \text{current parameter-group gradient} \\
        S_{t-1,1:H} & \text{history slots from step }t-1 \\
        c_{t-1,2:H} & \text{slot accumulation counters from step }t-1 \\
        H, r & \text{history depth and accumulation factor}
    \end{array}$
    \ENSURE $S_{t,1:H}, c_{t,2:H}$
    \STATE $(S_{t,1:H},c_{t,2:H}) \leftarrow (S_{t-1,1:H},c_{t-1,2:H})$
    \STATE $\mathrm{carry}\leftarrow S_{t-1,1}$; $S_{t,1}\leftarrow g_t$
    \hfill\algorithmiccomment{insert the current gradient}
    \STATE $h\leftarrow2$
    \WHILE{$h\le H$ and $\mathrm{carry}\neq\mathrm{empty}$}
        \IF{$c_{t-1,h}=0$}
            \STATE $\mathrm{next}\leftarrow\mathrm{empty}$; $S_{t,h}\leftarrow\mathrm{carry}$; $c_{t,h}\leftarrow1$
            \hfill\algorithmiccomment{fill an empty slot}
        \ELSIF{$c_{t-1,h}=r$}
            \STATE $\mathrm{next}\leftarrow S_{t-1,h}$; $S_{t,h}\leftarrow\mathrm{carry}$; $c_{t,h}\leftarrow1$
            \hfill\algorithmiccomment{replace the full slot and carry its value}
        \ELSE
            \STATE $\mathrm{next}\leftarrow\mathrm{empty}$
            \STATE $S_{t,h}\leftarrow\bigl(c_{t-1,h}S_{t-1,h}+\mathrm{carry}\bigr)/(c_{t-1,h}+1)$
            \STATE $c_{t,h}\leftarrow c_{t-1,h}+1$
            \hfill\algorithmiccomment{merge by averaging}
        \ENDIF
        \STATE $\mathrm{carry}\leftarrow\mathrm{next}$; $h\leftarrow h+1$
    \ENDWHILE
    \STATE \textbf{return} $S_{t,1:H}, c_{t,2:H}$
\end{algorithmic}
\end{algorithm}

\subsection{Optimizer-training tasks and protocol}

\begin{table}[H]
    \centering
    \scriptsize
    \caption{Optimizer-training tasks. Parameter counts include all trainable parameters.}
    \label{tab:tasks}
    \setlength{\tabcolsep}{2.5pt}
    \begin{tabularx}{\linewidth}{@{}ll>{\raggedright\arraybackslash}Xr@{}}
        \hline
        Model & Dataset & Architecture & Params. \\
        \hline
        MLP & Synthetic binary & 64 inputs; two 64-unit ReLU layers; 2-class head & $8{,}450$ \\
        ConvNet & CIFAR-10 & $5\!\times\!5$ convolutions with 2, 4 channels; ReLU and $2\!\times\!2$ pooling; 16-unit ReLU; 10-class head & $2{,}142$ \\
        Residual ConvNet & Fashion-MNIST & 8-channel $3\!\times\!3$ convolution with batch normalization; residual blocks of width 8, 16; adaptive pooling; 10-class head & $5{,}122$ \\
        ViT & CIFAR-10 & $8\!\times\!8$ patches; width 16; 2 heads; one pre-normalized block; feed-forward width 64; 10-class head & $6{,}810$ \\
        \hline
    \end{tabularx}
\end{table}

CIFAR-10 and Fashion-MNIST inputs are normalized channelwise with mean and standard deviation
0.5, without augmentation. The single-block ViT additionally uses a learned class token, learned positional
embeddings, and zero dropout. The synthetic dataset contains 20,000 standardized 64-dimensional
examples with 32 informative and 10 redundant features and class separation 0.8. Each training
set is split stratified 80/20 into model-training and held-out meta-objective partitions.

\subsubsection{Normalized held-out-loss objective}
\label{app:meta-objective}

The complete 20\% held-out partition is evaluated after every simulated update in microbatches of
512. The denominator $L_t^{\mathrm{held}}$ is detached and lower-bounded by $0.1$.
Contributions are averaged across the $K$ updates and four optimizer-training tasks.
Feature statistics are also detached; gradients pass through every other optimizer path, simulated
update, and subsequent held-out loss.
Table~\ref{tab:meta-training-config} summarizes the remaining optimizer-training configuration.

\begin{table}[H]
    \centering
    \small
    \caption{Optimizer-training configuration.}
    \label{tab:meta-training-config}
    \begin{tabularx}{\linewidth}{p{0.32\linewidth}X}
        \hline
        Item & Value \\
        \hline
        Tasks per optimizer-training update & All four tasks in Table~\ref{tab:tasks} \\
        Model-training batch & 128 \\
        Reset trajectories / steps per trajectory & 4 / 160 \\
        Maximum optimizer-training updates & 64 \\
        Truncated backpropagation horizon & 20 model-training updates \\
        Optimizer-training update interval & 10 model-training updates \\
        Meta-objective & Mean normalized held-out loss over $K$ updates \\
        Meta-gradient & Differentiation through simulated updates and held-out losses \\
        Meta-optimizer & Adam, learning rate $10^{-3}$, $\beta_1=0.9$, $\beta_2=0.999$,
        $\epsilon=10^{-8}$, global gradient-norm clipping at 10 \\
        Held-out evaluation microbatch & 512 \\
        Objective schedule & Model-training task loss only \\
        \hline
    \end{tabularx}
\end{table}

\subsection{Target-model training and evaluation}

\subsubsection{Baseline protocols}
The fixed-reference comparison uses Adam with learning rate $10^{-3}$,
$(\beta_1,\beta_2)=(0.9,0.999)$, $\epsilon=10^{-8}$, zero weight decay,
and no learning-rate schedule.

The task-tuned comparison selects Adam, AdamW, SGDM, or AdEMAMix hyperparameters by
the lowest validation loss for each target. The learned optimizer is
not adapted to any target task. VeLO, Celo2, and $\mu$LO use pretrained optimizer
networks without target-specific optimizer training; Celo2 follows its authors'
recommended deployment configuration.
Runtime profiles use CUDA implementations of Celo2 and VeLO verified against
their unfused implementations.

Adam learning rates are selected from
$\{10^{-4},3\!\times\!10^{-4},10^{-3},3\!\times\!10^{-3},10^{-2}\}$ for
BERT-Tiny, $\{3\!\times\!10^{-4},10^{-3},3\!\times\!10^{-3},10^{-2}\}$ for
GPT-Tiny, $\{10^{-4},3\!\times\!10^{-4},10^{-3},3\!\times\!10^{-3}\}$ for
ViT, and $\{3\!\times\!10^{-4},10^{-3},3\!\times\!10^{-3}\}$ separately for
each GNN model--dataset pair. SGDM learning rates are selected from
$\{3\!\times\!10^{-3},10^{-2},3\!\times\!10^{-2},10^{-1},3\!\times\!10^{-1}\}$ for
the language targets, $\{3\!\times\!10^{-3},10^{-2},3\!\times\!10^{-2}\}$ for GNNs,
and $\{3\!\times\!10^{-3},10^{-2},3\!\times\!10^{-2},10^{-1}\}$ for ViT.
AdEMAMix uses $\alpha=10$ and $\beta_3=0.9999$ on language targets,
screening $\{3\!\times\!10^{-4},10^{-3},3\!\times\!10^{-3}\}$.
The GNN comparison uses $\alpha=2$, $\beta_3=0.9999$, and the same learning-rate grid.
On ViT, the four candidates combine $\alpha\in\{5,10\}$ and
$\beta_3\in\{0.999,0.9999\}$, a subset of the values tested by
\citet[Appendix~B.3]{pagliardini2024ademamix}, with learning rate fixed at
$3\!\times\!10^{-4}$.
Learning-rate decay and weight averaging are not applied in
these classical-baseline comparisons.

\paragraph{AdamW search.}
AdamW learning rate and weight decay are selected jointly from the following combinations:
\begin{itemize}
    \item BERT-Tiny: learning rates $\{3\!\times\!10^{-4},10^{-3},3\!\times\!10^{-3}\}$
    crossed with weight decays $\{0.001,0.01,0.1\}$ (9 candidates).
    \item GPT-Tiny: learning rates $\{10^{-4},3\!\times\!10^{-4},10^{-3},3\!\times\!10^{-3}\}$
    crossed with weight decays $\{0.01,0.05,0.1\}$ (12 candidates).
    \item ViT: learning rates $\{10^{-4},3\!\times\!10^{-4},10^{-3},3\!\times\!10^{-3}\}$
    crossed with weight decays $\{0.01,0.05,0.1\}$, plus learning rate $3\!\times\!10^{-5}$
    with weight decays $\{0.05,0.1,0.3\}$ and learning rates $\{10^{-4},3\!\times\!10^{-4}\}$
    with weight decay $0.3$ (17 candidates).
    \item Each GNN model--dataset pair: learning rates
    $\{3\!\times\!10^{-4},10^{-3},3\!\times\!10^{-3},10^{-2}\}$
    crossed with weight decays $\{0.0005,0.005,0.05\}$ (12 candidates).
\end{itemize}
AdamW uses $(\beta_1,\beta_2)=(0.9,0.999)$ and $\epsilon=10^{-8}$.
Weight decay applies to tensors with at least two dimensions, excluding bias and normalization parameters.
Selection uses terminal validation loss for language models and the lowest validation loss during
training for ViT and GNNs.

\paragraph{Celo2 configuration.}
Language targets use the published LM-30M recipe with learning rate $4.64\times10^{-4}$,
weight decay $0.1$, linear warmup over the first $5\%$ of updates, and cosine decay to zero.
AdamW updates embeddings, output heads, biases, and one-dimensional tensors.
Celo2 updates the other matrices with orthogonalization enabled.
ViT and GNNs use a constant learning rate of $10^{-3}$ and zero weight decay,
with Celo2 applied to tensors of two or more dimensions and AdamW to one-dimensional tensors.
Orthogonalization is enabled in both cases. The AdamW branches use
$(\beta_1,\beta_2)=(0.9,0.95)$ and $\epsilon=10^{-8}$.

\paragraph{$\mu$LO configuration.}
The released $\mu$LO$_M$ weights are held fixed and applied to all target parameters
using the native PyTorch implementation~\citep{therien2024mu}.
The external learning-rate multiplier is 1, weight decay is zero, and gradient
clipping is disabled. The update and exponential multipliers are 0.01 and 0.001.
ViT follows the authors' $\mu$P initialization and scaling rules.
For language models, these rules are adapted to preserve shared embedding and
output-head weights and the original parameter counts.
For GCN, GraphSAGE, and GAT, we extend the input and output scaling rules to graph
layers, with output projections scaled by the inverse hidden width.
No target hyperparameter search or optimizer retraining is performed.

\subsubsection{Language models}

Both language targets use two layers, width 128, and two heads. BERT-Tiny follows
\citet{turc2019wellread}. The causal target uses a GPT-2 decoder~\citep{radford2019language}.
Table~\ref{tab:language-configs} summarizes their training configurations.

\begin{table}[H]
    \centering
    \small
    \caption{Language-model configurations used for optimizer evaluation.}
    \label{tab:language-configs}
    \begin{tabularx}{\linewidth}{p{0.14\linewidth}p{0.13\linewidth}p{0.16\linewidth}p{0.12\linewidth}X}
        \hline
        Model & Layers / width / heads & Context & Batch / microbatch & Data and objective \\
        \hline
        BERT-Tiny & 2/128/2 & 128 for 90k, then 512 for 10k & 256/256 & Wikipedia--BookCorpus-compatible masked-token prediction \\
        GPT-Tiny & 2/128/2 & 128 fixed & 64/64 & WikiText-103 next-token prediction \\
        \hline
    \end{tabularx}
\end{table}

The learned optimizer either updates all language-model parameters or assigns embeddings
and the output head to Adam, as indicated in each table and figure.
Language AdEMAMix comparisons also assign embeddings and the output head to Adam at learning rate $10^{-3}$.
Paired runs share initialization, data, stochastic streams, and validation data consisting of 20 microbatches.
Validation restores model mode and random-number state, and loss is aggregated over predicted
targets.

\subsubsection{Vision-transformer protocol}
\label{app:vit-protocol}

The CIFAR-100 model uses six Transformer blocks, width 256, eight heads, $8\!\times\!8$ patches, and dropout 0.1. The learned optimizer updates all parameters. 
A fixed stratified split supplies 45,000 training and 5,000 validation examples, with batch size 64 and 704 updates per epoch. 
Validation is measured after each of 50 epochs. Test accuracy is evaluated at the checkpoint
with the lowest validation loss. Terminal test accuracy, where reported, is evaluated separately
after epoch 50.

\subsubsection{Graph models}
The graph evaluation crosses GCN, GraphSAGE, and GAT with Cora, CiteSeer, and PubMed. Cora and CiteSeer run for 100 updates with seven slots. PubMed runs for 500 updates with nine slots. 

The public Planetoid split and stored node features are used without input-feature normalization.
Each split has 20 labeled training nodes per class, 500 validation nodes, and 1,000 test nodes, giving 140, 120, and 60 training nodes for Cora, CiteSeer, and PubMed. 
Training loss uses the training mask. 
Full-graph passes retain all nodes and edges. 
Validation is recorded after every update. The checkpoint with the lowest
validation loss is evaluated once on the test mask.

\subsection{Resource measurements}
\label{app:resource-measurements}

\begin{figure*}[t]
    \centering
    \includegraphics[width=\textwidth]{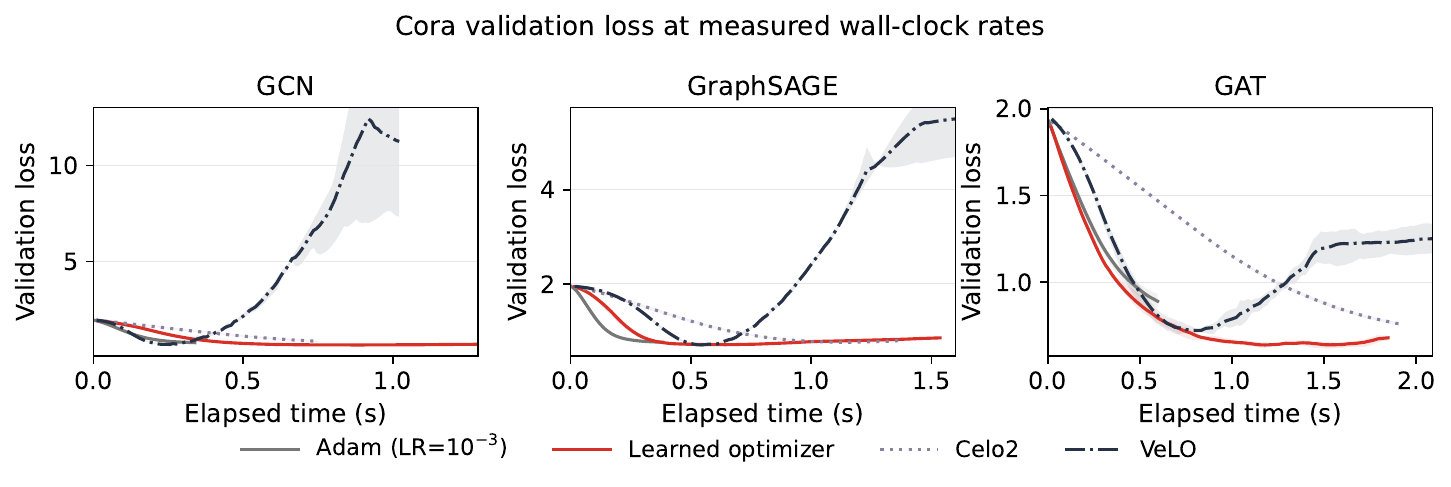}
    \caption{Cora validation loss against training time calculated from measured step times for all three graph architectures. The presented optimizer uses seven history slots covering the full trajectory. The gray Adam curves end after 100 updates. Each trajectory retains its complete recorded update budget, so slower optimizers extend farther along the time axis. Lines are three-seed means, and bands are pointwise observed minimum--maximum ranges. Celo2 and VeLO use the verified CUDA implementations.}
    \label{fig:gnn-time-matched-complete}
\end{figure*}

The reported $37{,}389$ parameters include trainable optimizer tensors and exclude fixed positional encodings.
All profiles use one NVIDIA B200, with the same model, batch, and parameter assignment within each comparison. Language profiling uses bfloat16 mixed precision, while ViT and GNN profiling use FP32.
Time/step measures a complete training step with CUDA events. External optimizers and their paired Adam controls run in fresh processes. Initialization, checkpoint loading, and CUDA compilation occur before timing. Each profile uses a warm-up followed by three measured blocks.
Inputs, gradients, and mature history state are resident on-device before timing. Data loading, validation, logging, and checkpointing are excluded. The GNN time ratios are geometric means across the nine model--dataset cells.
Persistent state is measured after all history slots are populated and includes learned-optimizer state, auxiliary Adam state, and optimizer-network
parameters.
Figure~\ref{fig:gnn-time-matched-complete} compares Cora validation loss against training time calculated from these measurements.

%% file: sections/B_additional_ablations.tex
\section{Additional Evaluation Results and Ablations}
\label{app:additional-results}

\subsection{Additional target-model results}

\subsubsection{Update-aligned language trajectories}

\begin{figure*}[t]
    \centering
    \includegraphics[width=0.96\textwidth]{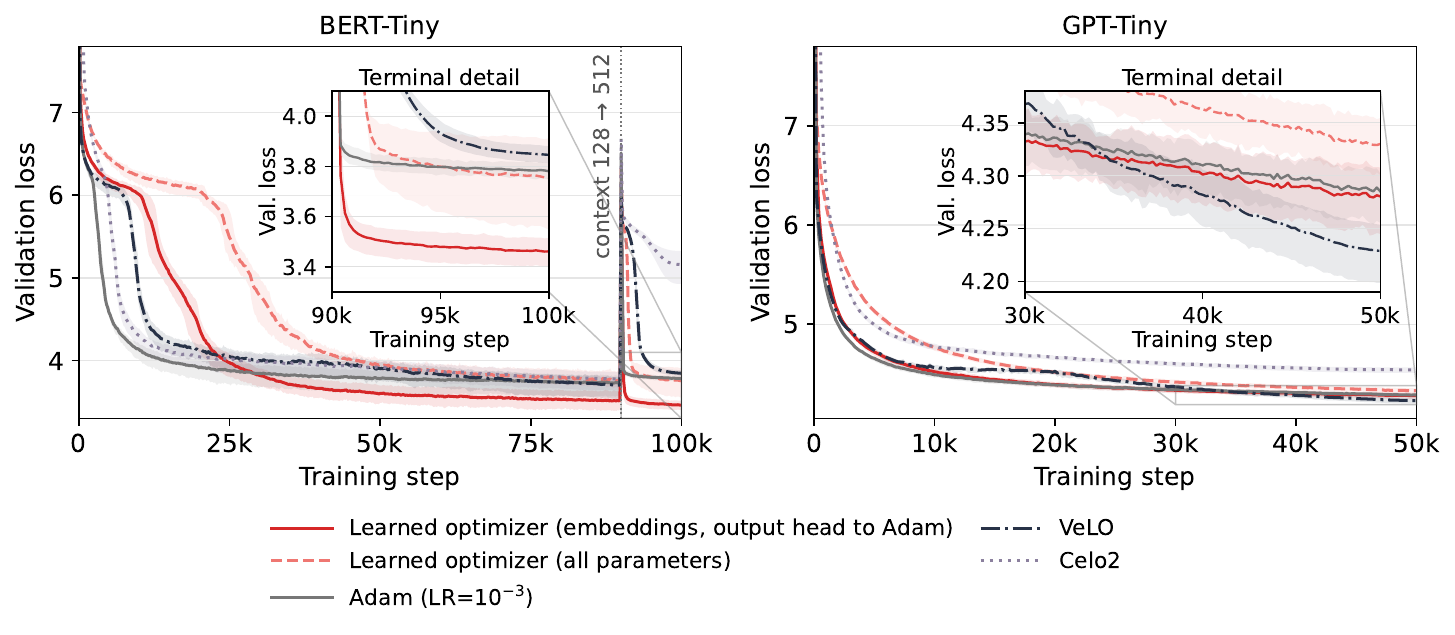}
    \caption{Update-aligned language-model validation loss. Lines show three-seed means, and bands
    show pointwise minimum--maximum ranges. Solid red assigns embedding and output-head tensors to
    Adam; dashed red applies the learned optimizer to all trainable tensors. BERT-Tiny changes
    context at update 90,000, and the GPT-Tiny inset magnifies the terminal trajectory.}
    \label{fig:language-trajectories-update-aligned}
\end{figure*}

Figure~\ref{fig:language-trajectories-update-aligned} separates optimization progress per update
from the elapsed-time comparison in Figure~\ref{fig:time-matched-language}.

\subsubsection{Longer language trajectories and a larger model}
\label{app:extended-language}

\begin{figure*}[t]
    \centering
    \includegraphics[width=0.96\textwidth]{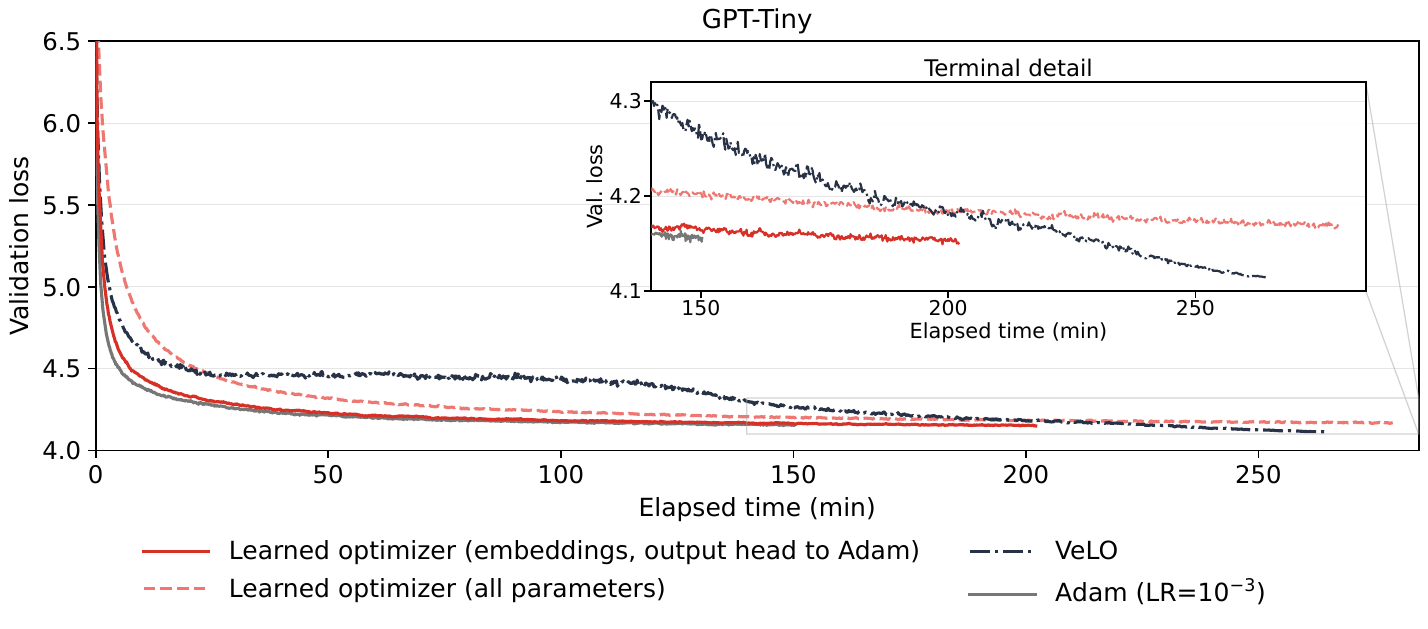}
    \caption{GPT-Tiny validation loss over 250k updates for one shared target seed.
    Elapsed time is calculated from measured step times. VeLO uses a planned horizon of 250k updates.}
    \label{fig:gpt-250k-time-matched}
\end{figure*}

At 250k updates, both Tiny-model parameter assignments remain finite across three seeds, and the
all-parameter BERT-Tiny result outperforms Adam in two of three seeds (Table~\ref{tab:extended-language}).
Figure~\ref{fig:gpt-250k-time-matched} extends the comparison with VeLO to 250k updates on GPT-Tiny.

\paragraph{Protocol.}
The Tiny models use 18 history slots over 250k updates, compared with six slots and 160-step
trajectories during optimizer training. BERT-Tiny changes from 128 to 512 tokens at update
225k, while GPT-Tiny keeps 128 tokens. BERT-Base uses 17 slots and changes context at update
90k within its 100k-update budget. It has 12 Transformer layers, width 768, and 12 attention heads.
Both BERT models use the Wikipedia--BookCorpus data with a 50M-token limit and masked-token
prediction without next-sentence prediction. GPT-Tiny uses WikiText-103.
BERT uses an effective batch of 256 sequences: one batch for Tiny and 16 microbatches of 16
for Base. GPT-Tiny uses one batch of 64 sequences. Model computation uses bfloat16 and optimizer
state uses float32.
Adam uses LR $10^{-3}$ and $\beta=(0.9,0.999)$ both as a baseline and when updating
the embedding and output-head tensors.

\subsubsection{Graph validation trajectories}
\label{app:gnn-training-trajectories}

\begin{figure*}[t]
    \centering
    \includegraphics[width=0.96\textwidth]{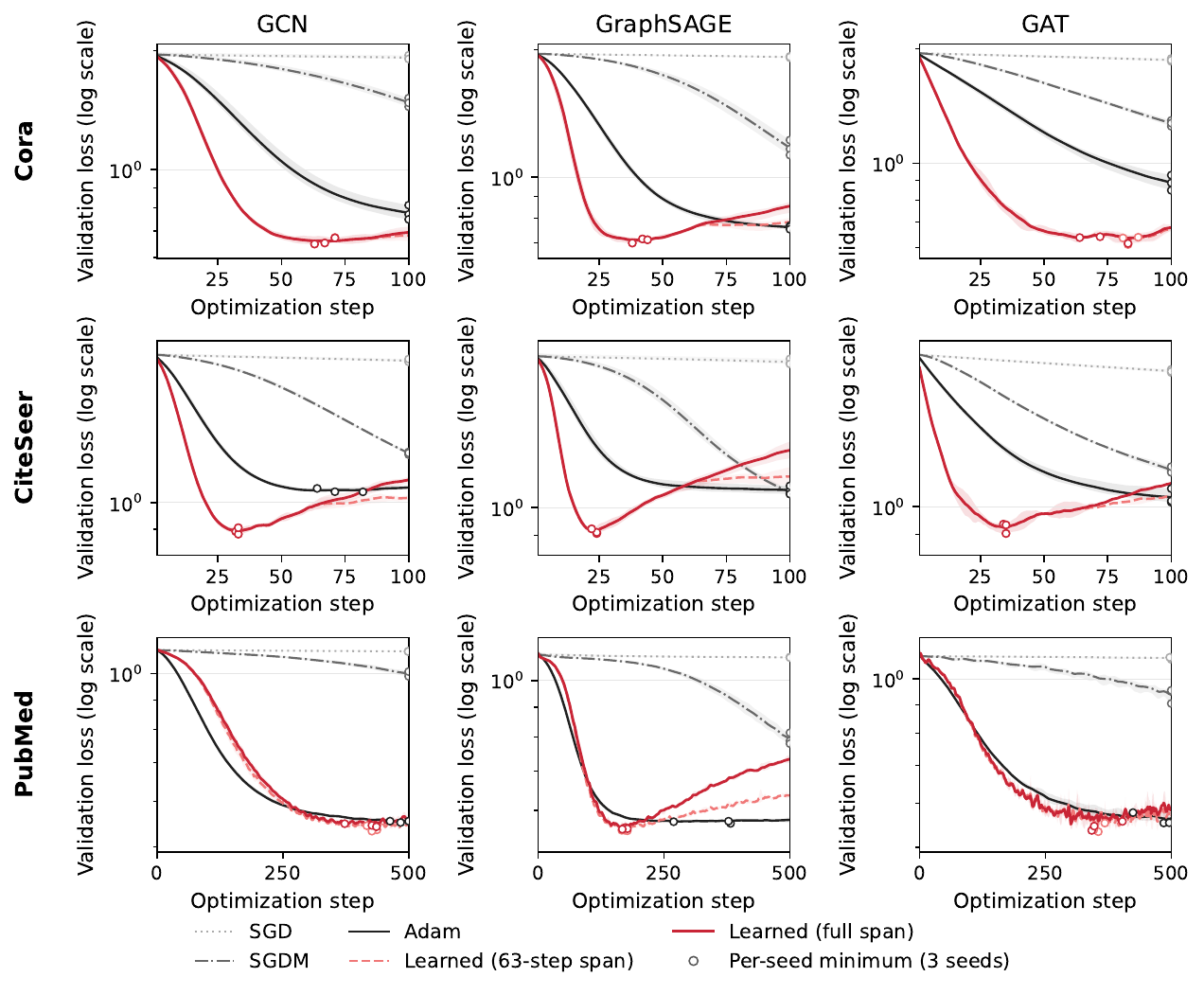}
    \caption{Step-aligned GNN validation loss. Rows identify datasets, and columns identify model
    architectures. Curves show three-seed means, bands show seedwise ranges, and circles mark the
    minimum of each seedwise validation trajectory. Checkpoints are selected independently within
    the fixed 100-update Cora and CiteSeer budgets and 500-update PubMed budget. Test accuracy does
    not enter checkpoint selection.}
    \label{fig:gnn-validation-trajectories}
\end{figure*}

Figure~\ref{fig:gnn-validation-trajectories} shows the complete validation signal used to select checkpoints without reference to test accuracy.

\subsection{Mechanism evidence and ablations}

\subsubsection{Parameter grouping}
\label{app:parameter-grouping}

Table~\ref{tab:parameter-grouping} compares changing parameter grouping at deployment with
training the optimizer for each grouping. Every trajectory remains finite, but effects vary
by task and metric. Each optimizer uses one optimizer-training seed, and the results do not identify
a preferred grouping.

\begin{table*}[h]
    \centering
    \small
    \caption{Parameter-grouping ablation. Entries are mean $\pm$ sample standard deviation over
    three paired target seeds. The group count is measured at deployment. Output and input slices
    prioritize complete leading- and second-dimension slices, respectively. Tensor-wide and
    network-wide conditions form one group per parameter tensor and target model. Bold marks the best
    mean within each task and metric. Adam uses the same target initializations and data streams.}
    \label{tab:parameter-grouping}
    \setlength{\tabcolsep}{3.5pt}
    \begin{tabular*}{\textwidth}{@{\extracolsep{\fill}}llrrr}
        \hline
        \shortstack[l]{Optimizer-training\\grouping} & \shortstack[l]{Deployment\\grouping} &
        \shortstack[r]{Deployment\\groups} & \shortstack[r]{Train-loss\\AUC $\downarrow$} & Terminal \\
        \hline
        \input{figures/data/TableB_ParameterGroupingRows.tex}
    \end{tabular*}
\end{table*}

\subsubsection{History-depth controls}
\label{app:history-depth}

Table~\ref{tab:gnn-history-depth} compares the 63-step span with the full span over three paired
target seeds. Aggregate accuracy is similar, while the preferred span varies across cells. History
depth is fixed before validation and test evaluation.

\begin{table}[H]
    \centering
    \footnotesize
    \caption{GNN history-depth ablation. Test accuracy (\%) at the minimum-validation-loss checkpoint,
    mean $\pm$ sample SD over three paired target seeds.
    Final row: unweighted mean across nine tasks. Bold and underline mark the best and second-best means.}
    \label{tab:gnn-history-depth}
    \setlength{\tabcolsep}{4pt}
    \begin{tabular*}{\linewidth}{@{\extracolsep{\fill}}llrr}
        \hline
        Data & Model & 63-step span (\%) & Full span (\%) \\
        \hline
        \input{figures/data/Table3_GNNDepthRows.tex}
    \end{tabular*}
\end{table}

\subsubsection{Temporal coefficients and stored-history content}
\label{app:temporal-dynamics}
\label{app:history-content}

\begin{figure*}[t]
    \centering
    \includegraphics[width=\textwidth]{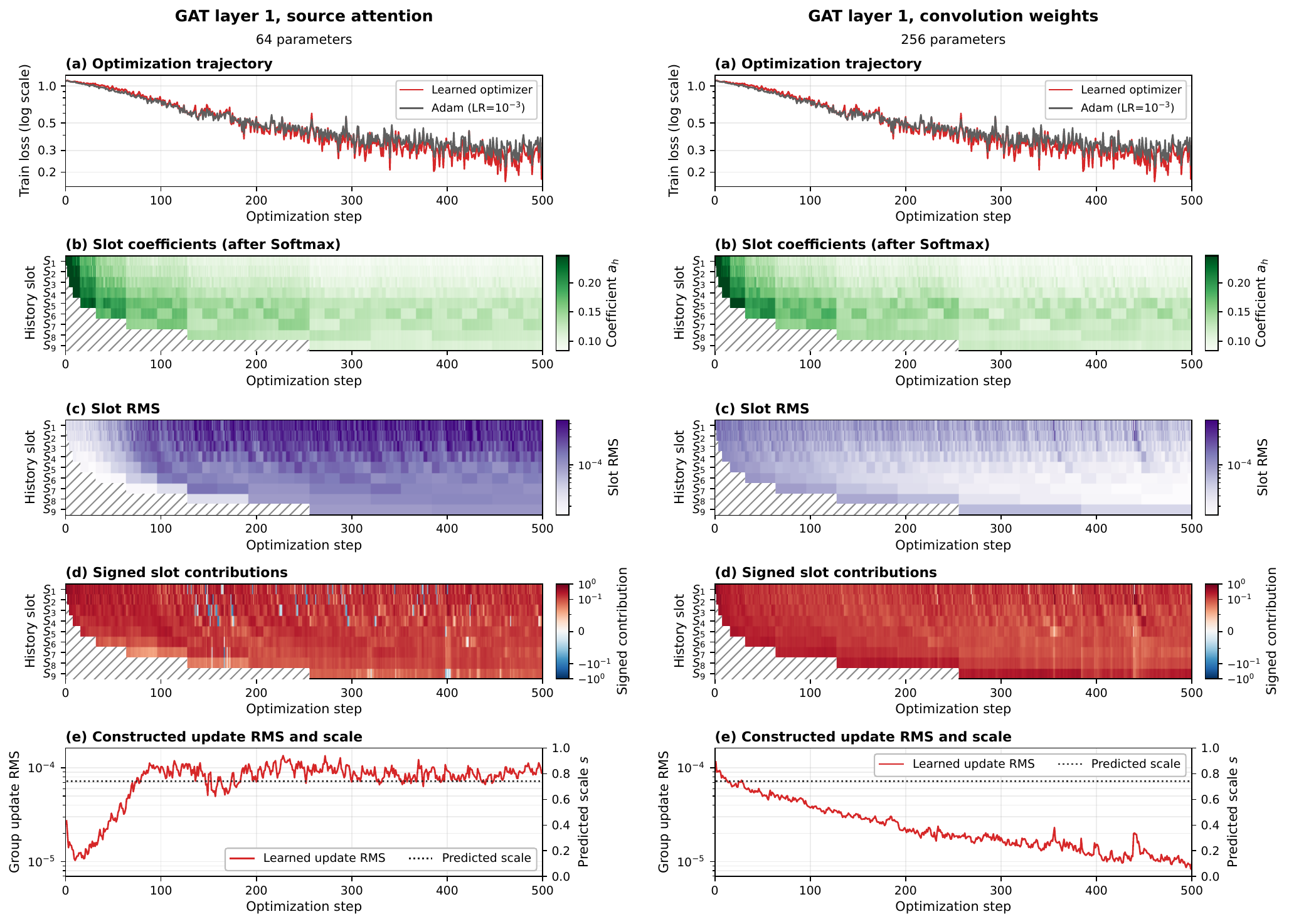}
    \caption{\textbf{Update norms and temporal dynamics on GAT/PubMed.} Columns show a
    64-coordinate source-attention group and a 256-coordinate convolution-weight group from layer 1
    over the same 500 updates. (a) Training loss. (b) Slot coefficients $a_h$ after SoftMax.
    (c) RMS of each stored gradient average. (d) Signed slot contributions
    $\langle u_h,u\rangle/\|u\|^2$, where $u_h$ is the update from slot $h$ and $u=\sum_h u_h$.
    (e) RMS of the constructed group update and predicted scale $s$.
    $S_1$ is the newest slot, and hatching marks unfilled slots.
    The convolution-weight group has the largest mean coefficient difference from the source-attention
    group among 126 weight groups.}
    \label{fig:gat-pubmed-long-temporal-scores}
\end{figure*}

The update magnitude reflects both stored gradient averages and their coefficients, rather than
the scale factor alone (Figure~\ref{fig:gat-pubmed-long-temporal-scores}). For the source-attention
group, update RMS varies 13.1-fold while the predicted scale ranges only from 0.74220 to 0.74257.
The convolution-weight group shows a different update-magnitude trajectory within the same model.

\subsubsection{Learned coefficients, scale, and self-attention}
\label{app:coefficient-scale-controls}

\begin{figure}[!p]
    \centering
    \includegraphics[width=\textwidth]{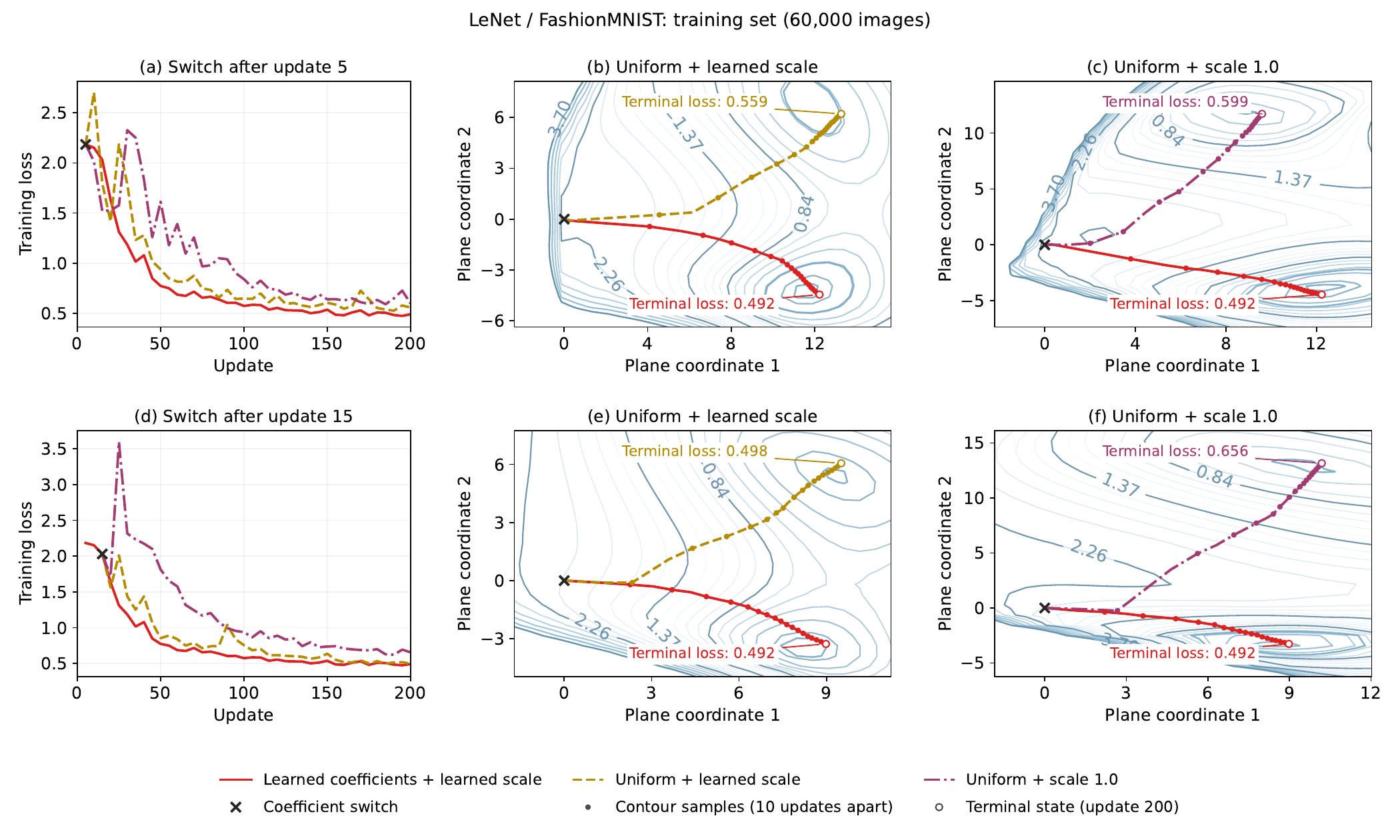}
    \caption{Coefficient switches after updates 5 and 15 on LeNet/Fashion-MNIST. Curves and contours evaluate loss on all 60,000 training images. Uniform coefficients retain the learned scale or fix it to 1 after the switch. Contours use a separate plane for each comparison, with trajectories projected onto that plane.}
    \label{fig:uniform-coefficient-branches-both}
    \vspace{10pt}
    \includegraphics[width=\textwidth]{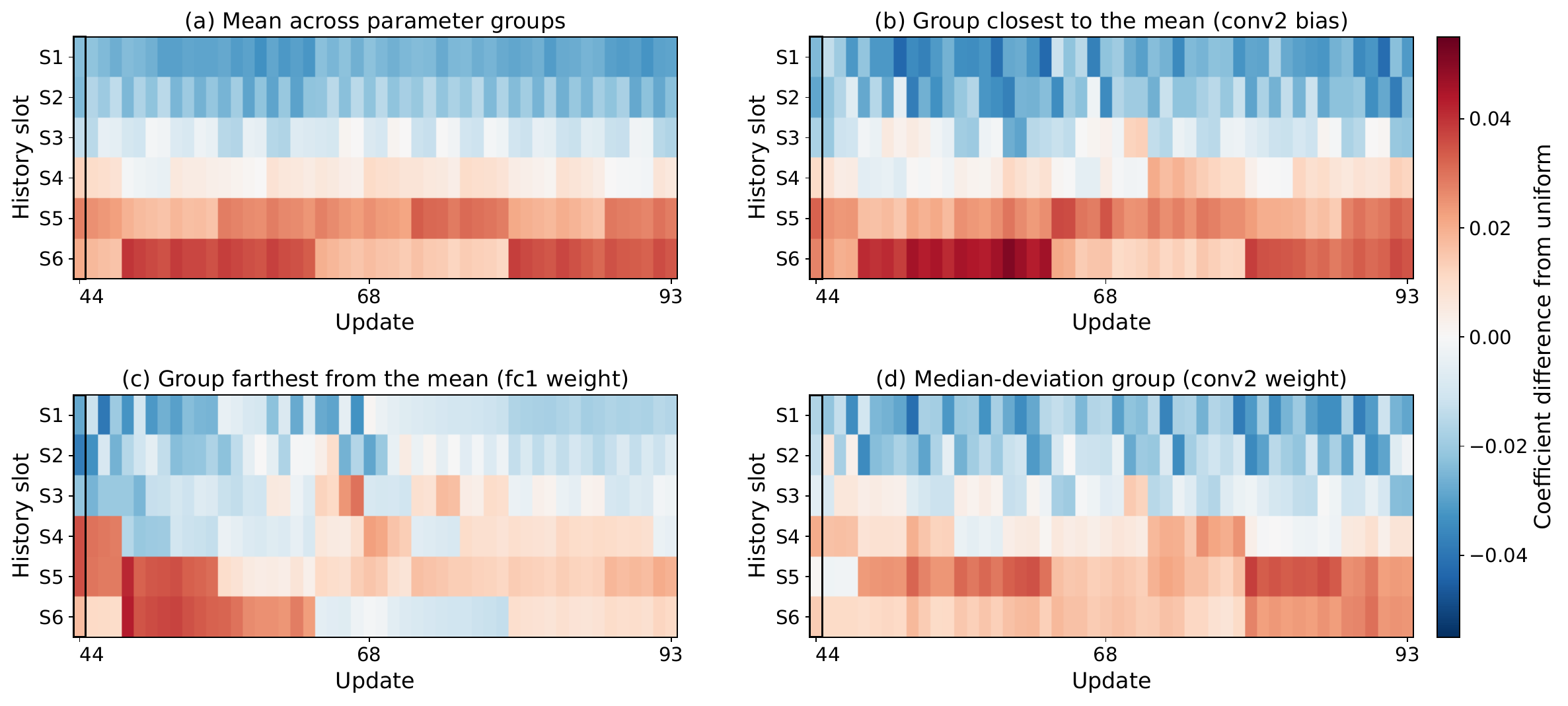}
    \caption{\textbf{History weighting varies across parameter groups.}
    All panels show temporal coefficient differences from uniform weighting at the same learned scale over updates 44--93 on the trajectory in Figure~\ref{fig:uniform-coefficient-branches}.
    (a) Equal-weight mean across all 188 parameter groups.
    (b) Group closest to the mean: conv2 bias, also shown in Figure~\ref{fig:uniform-coefficient-branches}(g).
    (c) Group farthest from the mean: a group from fc1 weights.
    (d) Median-deviation group (conv2 weight), a middle-ranked example between (b) and (c).
    Selection is defined in Appendix~\ref{app:group-coefficient-variation}.
    The color scale matches Figure~\ref{fig:uniform-coefficient-branches}(f--g).
    The outlined column marks the coefficient switch at update 44, and slots run newest to oldest.}
    \label{fig:group-coefficient-variation}
\end{figure}

Switching from learned coefficients to uniform coefficients worsens descent in the LeNet experiment, and fixing the scale to 1 further increases accumulated training loss.
Figure~\ref{fig:uniform-coefficient-branches-both} shows the loss trajectories and contours for coefficient switches after updates 5 and 15.
From each shared model and history state, we switch from learned coefficients to uniform coefficients over populated slots, retaining the learned scale or fixing it to 1.
Table~\ref{tab:coefficient-branch-auc} integrates full-training-set loss from the coefficient switch through update 200 using the trapezoidal rule.

\begin{table}[h]
    \centering
    \small
    \caption{Increase in training-loss AUC relative to the unchanged learned optimizer (\%, $\downarrow$).}
    \label{tab:coefficient-branch-auc}
    \renewcommand{\arraystretch}{1.2}
    \begin{tabular*}{\textwidth}{@{\extracolsep{\fill}}lrrr}
        \hline
        \shortstack[l]{Update before\\coefficient switch} & \shortstack{Learned\\optimizer} & \shortstack{Uniform coefficients\\and learned scale} & \shortstack{Uniform coefficients\\and scale 1} \\
        \hline
        5 & 0.0 & +16.4 & +37.0 \\
        15 & 0.0 & +17.0 & +71.5 \\
        \hline
    \end{tabular*}
\end{table}

With scale fixed to 1 on ViT/CIFAR-100, all three learned-coefficient runs complete 50 epochs, while all three uniform-coefficient runs diverge with nonfinite gradients (Table~\ref{tab:vit-fixed-scale-stability}).

\begin{table}[h]
    \centering
    \small
    \caption{ViT stability with the scale fixed to 1 for both coefficient rules.}
    \label{tab:vit-fixed-scale-stability}
    \renewcommand{\arraystretch}{1.2}
    \begin{tabular*}{\textwidth}{@{\extracolsep{\fill}}lrr}
        \hline
        & \shortstack{Learned optimizer\\with scale 1} & \shortstack{Uniform coefficients\\with scale 1} \\
        \hline
        Runs with nonfinite gradients $\downarrow$ & \textbf{0/3} & 3/3 (all diverged) \\
        \hline
    \end{tabular*}
\end{table}

Removing self-attention reduces GAT/PubMed test accuracy by 29.77 percentage points (Table~\ref{tab:attention-pubmed}).
The optimizer without self-attention is retrained and evaluated on the same target seeds.

\begin{table}[h]
    \centering
    \small
    \caption{GAT/PubMed test accuracy (\%) at the checkpoint with the lowest validation loss, reported as mean $\pm$ sample standard deviation over three target seeds. Both optimizers use six history slots.}
    \label{tab:attention-pubmed}
    \renewcommand{\arraystretch}{1.2}
    \begin{tabular*}{\textwidth}{@{\extracolsep{\fill}}lrr}
        \hline
        & Learned optimizer & \shortstack{Learned optimizer\\without self-attention} \\
        \hline
        Test accuracy (\%) $\uparrow$ & 78.77 $\pm$ 0.21 & 49.00 $\pm$ 7.74 \\
        \hline
    \end{tabular*}
\end{table}

\subsubsection{Coefficient variation across parameter groups}\label{app:group-coefficient-variation}

Groups in Figure~\ref{fig:group-coefficient-variation} are ranked by the RMS difference between their temporal coefficients $\alpha$ and the equal-group mean across all six slots and updates 44--93.
Panels (b--d) show the closest, farthest, and median-deviation groups, respectively.

In Figure~\ref{fig:uniform-coefficient-branches}, the whole-network loss plane spans the learned and same-scale uniform endpoint displacements from the shared state at update 43.
Intermediate states and the scale-1 endpoint can lie outside that plane, so endpoint annotations and the loss curve use actual full-training-set evaluations.
Mean loss in panel (b) is trapezoidal training-loss AUC over updates 43--160, divided by 117.
At the coefficient switch, the orthogonal contribution axis points toward the same-scale uniform update's residual after removing its projection onto the learned update.

%% file: figures/data/TableB_ParameterGroupingRows.tex
\textit{LeNet/CIFAR-10 (terminal test loss $\downarrow$)} & & & & \\
Output slice & Output slice & 15 & \textbf{1.820 $\pm$ 0.012} & 1.743 $\pm$ 0.055 \\
Output slice & Input slice & 14 & 1.823 $\pm$ 0.016 & 1.705 $\pm$ 0.054 \\
Output slice & Tensor-wide & 8 & 1.822 $\pm$ 0.014 & 1.681 $\pm$ 0.028 \\
Output slice & Network-wide & 1 & 1.831 $\pm$ 0.017 & 1.666 $\pm$ 0.018 \\
\cline{1-5}
Input slice & Input slice & 14 & 1.873 $\pm$ 0.043 & 1.748 $\pm$ 0.034 \\
Tensor-wide & Tensor-wide & 8 & 1.827 $\pm$ 0.024 & \textbf{1.665 $\pm$ 0.034} \\
Network-wide & Network-wide & 1 & 1.844 $\pm$ 0.009 & 1.693 $\pm$ 0.017 \\
Adam & -- & -- & 1.972 $\pm$ 0.024 & 1.804 $\pm$ 0.008 \\
\hline
\textit{GAT/Cora (terminal test accuracy $\uparrow$)} & & & & \\
Output slice & Output slice & 367 & 0.195 $\pm$ 0.007 & 80.93 $\pm$ 0.15 \\
Output slice & Input slice & 367 & 0.195 $\pm$ 0.007 & \textbf{81.13 $\pm$ 0.31} \\
Output slice & Tensor-wide & 8 & 0.194 $\pm$ 0.007 & 81.00 $\pm$ 0.26 \\
Output slice & Network-wide & 1 & 0.194 $\pm$ 0.007 & 81.00 $\pm$ 0.10 \\
\cline{1-5}
Input slice & Input slice & 367 & \textbf{0.182 $\pm$ 0.007} & 80.90 $\pm$ 0.10 \\
Tensor-wide & Tensor-wide & 8 & 0.193 $\pm$ 0.007 & 81.00 $\pm$ 0.30 \\
Network-wide & Network-wide & 1 & 0.186 $\pm$ 0.007 & 81.00 $\pm$ 0.36 \\
Adam & -- & -- & 0.515 $\pm$ 0.013 & 80.37 $\pm$ 1.03 \\
\hline

%% file: figures/data/Table3_GNNDepthRows.tex
        Cora & GCN & \underline{81.67 $\pm$ 0.15} & \textbf{81.70 $\pm$ 0.20} \\
         & GraphSAGE & \textbf{79.60 $\pm$ 0.44} & \textbf{79.60 $\pm$ 0.44} \\
         & GAT & \textbf{81.97 $\pm$ 0.57} & \textbf{81.97 $\pm$ 0.31} \\
        \hline
        CiteSeer & GCN & \textbf{71.03 $\pm$ 0.81} & \textbf{71.03 $\pm$ 0.81} \\
         & GraphSAGE & \textbf{69.87 $\pm$ 0.87} & \textbf{69.87 $\pm$ 0.87} \\
         & GAT & \textbf{69.10 $\pm$ 0.26} & \underline{69.00 $\pm$ 0.26} \\
        \hline
        PubMed & GCN & \textbf{79.10 $\pm$ 0.17} & \underline{78.67 $\pm$ 0.21} \\
         & GraphSAGE & \underline{77.17 $\pm$ 0.12} & \textbf{77.37 $\pm$ 0.49} \\
         & GAT & \textbf{78.77 $\pm$ 0.21} & \underline{78.73 $\pm$ 0.31} \\
        \noalign{\hrule height 1.1pt}
        \multicolumn{2}{l}{Arithmetic mean} & \textbf{76.47 $\pm$ 0.24} & \underline{76.44 $\pm$ 0.28} \\
        \hline

%% file: sections/C_structural_extension.tex
\section{Spatio-Temporal Extension}
\label{app:spatio-temporal-extension}

This section extends temporal recombination across spatial tiers: multiple partitions of a target-model tensor independently construct candidate updates from gradient history, and a second predictor combines those candidates into the final parameter update.

\subsection{Pyramid grouping}

Pyramid grouping tests whether the learned optimizer can operate over more than one fixed
partition of a parameter tensor. Each tensor of rank at least two is represented by its first axis
and the flattened product of its remaining axes. 

\begin{figure*}[t]
    \centering
    \includegraphics[width=0.9\textwidth]{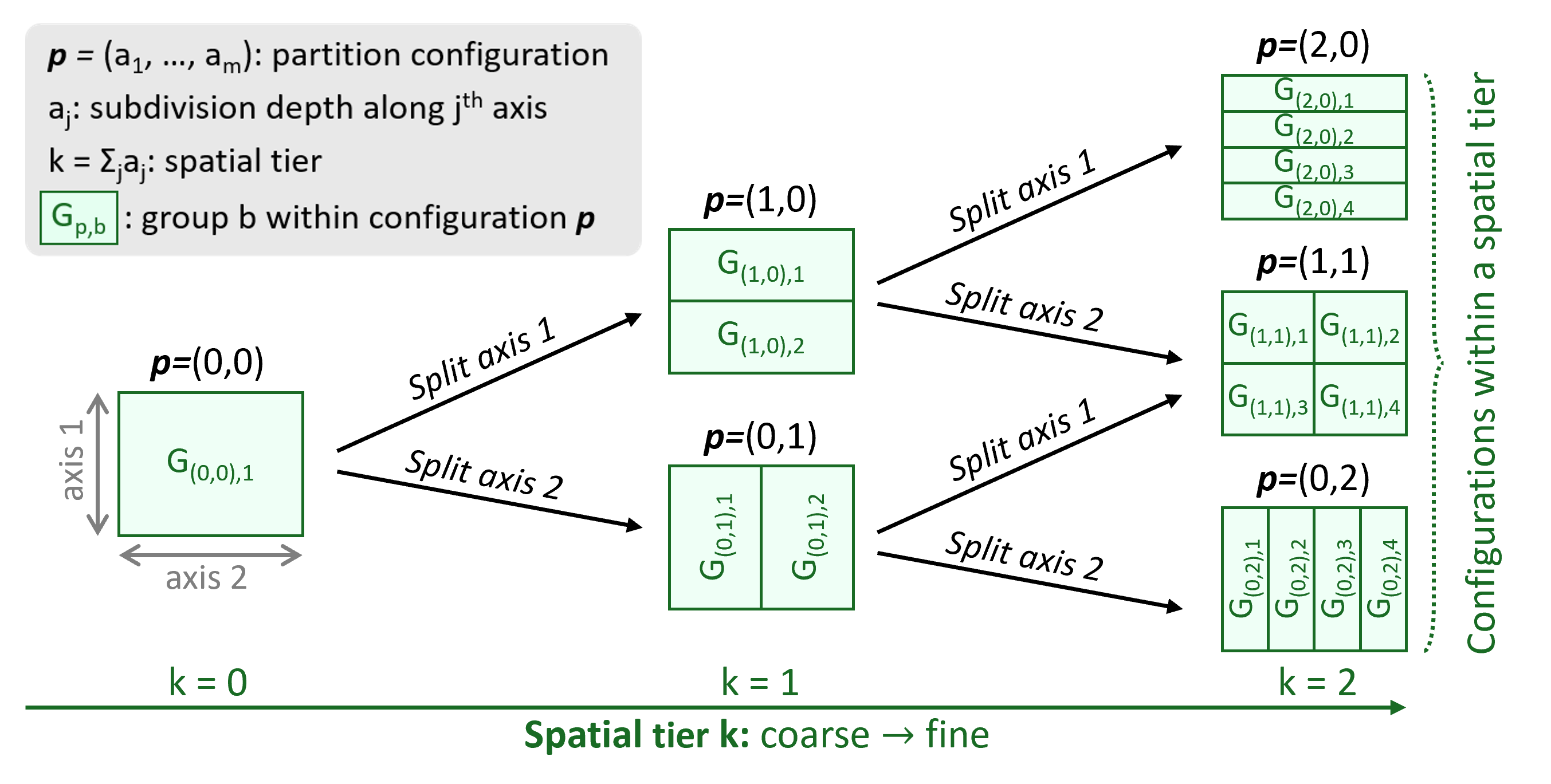}
    \caption{Spatio-temporal extension through pyramid grouping. A partition configuration
    $p=(a_1,\ldots,a_m)$ assigns subdivision depth $a_j$ to tensor axis $j$, and spatial tier
    $k=\sum_j a_j$ contains configurations with the same total subdivision depth. Each cell
    $G_{p,b}$ denotes parameter group $b$ within configuration $p$. The illustration uses two
    tensor axes and split factor $s=2$ through tier $k=2$; the evaluated realization retains the
    whole-tensor and two single-axis configurations with 16 splits per active axis. Within each
    group, the learned optimizer recombines the stored gradient history.}
    \label{fig:pyramid-partitioning}
\end{figure*}

Figure~\ref{fig:pyramid-partitioning} illustrates the generalized spatial hierarchy. The evaluated
realization retains the whole-tensor configuration and the two single-axis configurations.

The learned optimizer produces one full-tensor candidate $\Delta\theta_t^{(p)}$ from each retained
partition $p$. A separate predictor assigns Softmax weights $\beta_t^{(p)}$ to these candidates and
constructs
\begin{equation}
    \Delta\theta_t = \sum_p \beta_t^{(p)}\Delta\theta_t^{(p)}.
\end{equation}
The temporal path applies the same temporal predictor, with shared weights, to every retained partition. 
The spatial path uses a separate network, also shared across partitions, whose feature encoder, self-attention encoder, and coefficient head follow the temporal architecture but have independently learned parameters. 
The spatial network omits the scale token and group-scale output; its Softmax weights $\beta_t^{(p)}$ directly combine the partition candidates. 

The evaluated construction forms three partitions: the whole tensor, 16 contiguous groups along the first axis, and 16 contiguous groups along the flattened trailing axis. 
Groups smaller than 64 coordinates are omitted.
The optimizer with pyramid grouping matches the optimizer with row grouping across nine unseen graph tasks.
Table~\ref{tab:pyramid-gnn} reports test accuracy from the minimum-validation-loss state. 
Its arithmetic mean is 76.51\%, compared with 76.44\% for row grouping and 73.78\% for Adam.
The two optimizers were trained independently, so the comparison establishes that both grouping constructions function under the evaluated protocol rather than isolating the effect of grouping alone.

\begin{table*}[t]
    \centering
    \small
    \caption{Full-trajectory test accuracy (\%) for row and pyramid grouping on unseen graph tasks. Cora and CiteSeer use 100 updates and PubMed uses 500. 
    Entries are mean $\pm$ sample standard deviation over three target seeds. 
    Bold and underlined values mark the best and second-best mean in each row.
    }
    \label{tab:pyramid-gnn}
    \setlength{\tabcolsep}{2.7pt}
    \renewcommand{\arraystretch}{1.22}
    \begin{tabular*}{\textwidth}{@{\extracolsep{\fill}}llrrrrr}
        \hline
        Data & Model & Row & Pyramid & Adam & SGD & SGDM \\
        \hline
        Cora & GCN & \textbf{81.70 $\pm$ 0.20} & \underline{81.57 $\pm$ 0.12} & 79.27 $\pm$ 0.83 & 31.80 $\pm$ 10.01 & 74.93 $\pm$ 0.92 \\
             & GraphSAGE & \underline{79.60 $\pm$ 0.44} & \textbf{80.00 $\pm$ 0.69} & 78.73 $\pm$ 0.32 & 24.10 $\pm$ 3.58 & 76.63 $\pm$ 0.15 \\
             & GAT & \underline{81.97 $\pm$ 0.31} & \textbf{82.03 $\pm$ 0.50} & 78.27 $\pm$ 1.26 & 47.23 $\pm$ 8.00 & 78.97 $\pm$ 1.19 \\
        \hline
        CiteSeer & GCN & \textbf{71.03 $\pm$ 0.81} & \underline{70.97 $\pm$ 0.67} & 66.20 $\pm$ 1.61 & 33.00 $\pm$ 6.22 & 70.57 $\pm$ 0.71 \\
             & GraphSAGE & 69.87 $\pm$ 0.87 & \underline{70.13 $\pm$ 0.29} & 65.57 $\pm$ 1.36 & 34.67 $\pm$ 9.38 & \textbf{70.23 $\pm$ 0.38} \\
             & GAT & 69.00 $\pm$ 0.26 & \underline{69.17 $\pm$ 0.32} & 66.20 $\pm$ 0.82 & 52.87 $\pm$ 1.42 & \textbf{70.37 $\pm$ 0.78} \\
        \hline
        PubMed & GCN & \textbf{78.67 $\pm$ 0.21} & \underline{78.53 $\pm$ 0.60} & 77.50 $\pm$ 0.36 & 51.77 $\pm$ 1.15 & 67.67 $\pm$ 6.11 \\
             & GraphSAGE & \textbf{77.37 $\pm$ 0.49} & \underline{77.23 $\pm$ 0.29} & 75.17 $\pm$ 0.35 & 50.20 $\pm$ 13.67 & 74.70 $\pm$ 0.36 \\
             & GAT & \underline{78.73 $\pm$ 0.31} & \textbf{78.97 $\pm$ 0.12} & 77.10 $\pm$ 1.42 & 59.23 $\pm$ 5.95 & 73.30 $\pm$ 0.17 \\
        \noalign{\hrule height 1.1pt}
        \multicolumn{2}{l}{Arithmetic mean across 9 cells} & \underline{76.44 $\pm$ 0.28} & \textbf{76.51 $\pm$ 0.16} & 73.78 $\pm$ 0.26 & 42.76 $\pm$ 2.36 & 73.04 $\pm$ 0.75 \\
        \hline
    \end{tabular*}
\end{table*}